\documentclass{article} 
\usepackage[table]{xcolor}
\usepackage{iclr2025_conference,times}

\usepackage{amsmath,amsfonts,bm}

\def\eqref#1{equation~\ref{#1}}

\def\1{\bm{1}}

\def\vc{{\bm{c}}}

\DeclareMathAlphabet{\mathsfit}{\encodingdefault}{\sfdefault}{m}{sl}
\SetMathAlphabet{\mathsfit}{bold}{\encodingdefault}{\sfdefault}{bx}{n}

\usepackage[utf8]{inputenc} 
\usepackage[T1]{fontenc}    
\usepackage{hyperref}       
\usepackage{url}            
\usepackage{booktabs}       
\usepackage{amsfonts}       
\usepackage{nicefrac}       
\usepackage{microtype}      

\usepackage{graphicx}
\usepackage{amssymb}
\usepackage{wrapfig}
\usepackage{enumitem}
\usepackage{amsfonts}
\usepackage{amsmath}
\usepackage{algorithm}
\usepackage{algorithmic}
\usepackage{listings}
\usepackage[toc,title,page]{appendix}
\usepackage{graphicx}
\usepackage{multirow}
\usepackage{xspace}

\title{Improving Multi-modal Large Language Model through Boosting Vision Capabilities}

\author{
Yanpeng Sun$^{1,2}$ \qquad Huaxin Zhang$^{2,3}$ \qquad Qiang Chen$^2$ \qquad Xinyu Zhang$^2$ \qquad Nong Sang$^3$\\[2mm]
\qquad \textbf{Gang Zhang}$^2$ \qquad \textbf{Jingdong Wang$^{2*}$} \qquad  \textbf{Zechao Li$^1$}\thanks{Corresponding author.}\\[3mm]
$^1$Nanjing University of Science and Technology,\\[2mm]
$^2$Baidu VIS,\qquad $^3$Huazhong University of Science and Technology\\[3mm]
{\tt\small \{yanpeng\_sun, zechao.li\}@njust.edu.cn}
}

\iclrfinalcopy 
\begin{document}

\maketitle

~\vspace{-6mm}
\begin{abstract}

We focus on improving the visual understanding capability for boosting the vision-language models. We propose \textbf{Arcana}, a multiModal language model, which introduces two crucial techniques. First, we present Multimodal LoRA (MM-LoRA), a module designed to enhance the decoder. Unlike traditional language-driven decoders, MM-LoRA consists of two parallel LoRAs -- one for vision and one for language -- each with its own parameters. This disentangled parameters design allows for more specialized learning in each modality and better integration of multimodal information. Second, we introduce the Query Ladder adapter (QLadder) to improve the visual encoder. QLadder employs a learnable ``\textit{ladder}'' structure to deeply aggregates the intermediate representations from the frozen pretrained visual encoder (e.g., CLIP image encoder). This enables the model to learn new and informative visual features, as well as remaining the powerful capabilities of the pretrained visual encoder. These techniques collectively enhance Arcana's visual perception power, enabling it to leverage improved visual information for more accurate and contextually relevant outputs across various multimodal scenarios. Extensive experiments and ablation studies demonstrate the effectiveness and generalization capability of our Arcana. The code and re-annotated data are available at \url{https://arcana-project-page.github.io}.

\end{abstract}
~\vspace{-2.5em}
\section{Introduction}
\label{sec:introduction}
In recent years, multimodal large language models (MLLMs)~\cite{wang2023cogvlm,bai2023qwen,llava,ye2023mplug} have made significant advancements. 
These models amalgamate image representations into large language models (LLMs) through an adaptor~\cite{touvron2023llama,zheng2024judging}.
Various methods~\cite{dai2024instructblip,llava,wang2023cogvlm,dong2024internlm}
leverage the capabilities of the powerful LLM to excel in various multimodal tasks.

\begin{figure}
	\centering
	\includegraphics[width=\linewidth]{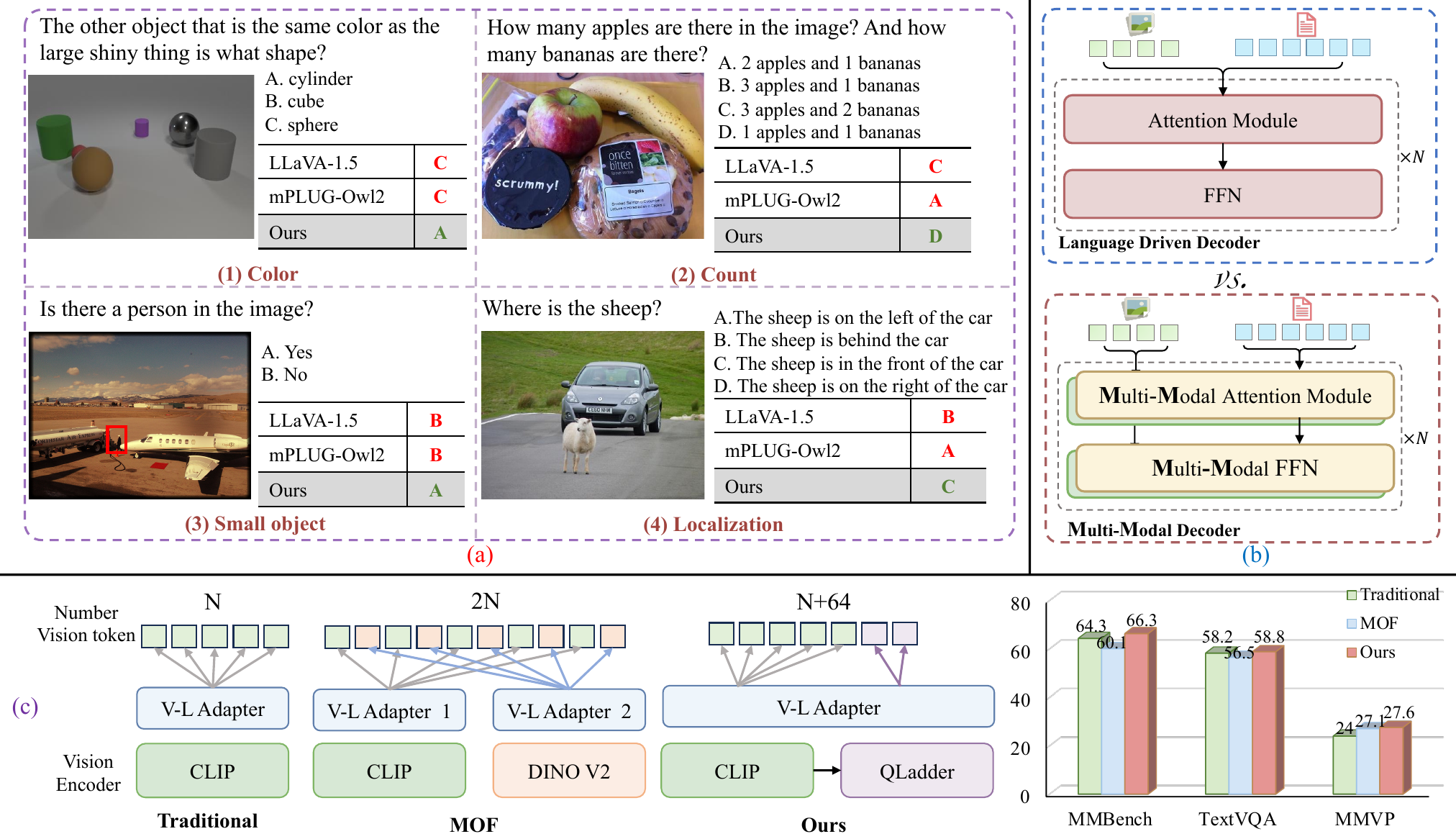}
	\vspace{-1.5em}
	\caption{\textbf{(a)} Sampled some VQA examples involving color, quantity, small objects, and localization tasks, showcasing the importance of visual recognition capabilities for multimodal language models (MLLMs). \textbf{(b)} Contrasting Arcana's multimodal decoder with mainstream methods' language driven decoder. The language-driven decoder employs a language decoder (LLMs) directly to handle tokens from different modalities, which may lead to modality interference and performance degradation. In contrast, the multimodal decoder independently processes different token types to avoid modality interference. \textbf{(c)} illustrates the structures of different visual encoders and the resulting number of visual tokens obtained. The bar chart displays the model's performance across various architectures.}
	\label{fig:motivation}
 \vspace{-1.6em}
\end{figure}

While existing MLLMs showcase remarkable proficiency in multimodal tasks, 
they still face challenges in visual perception
that is crucial for further tasks, such as reasoning or creation~\cite{chen2023shikra,liu2023llava1.5}.
Fig.~\ref{fig:motivation} (a) presents several examples that clearly highlight this issue. 
We observe deficiencies in current MLLMs regarding low-level visual perception, such as color and quantity, as well as high-level visual perception, such as small object detection and localization. Consequently, there is a pressing necessity to bolster the comprehension capabilities of existing MLLMs, specially for \emph{vision}.

The insufficient visual perception capabilities of MLLMs can mainly be attributed to two factors: \textit{decoder} and \textit{visual encoder}. 
As depicted in Fig.~\ref{fig:motivation}(b), existing language driven decoder structures directly couple visual and language modalities. 
Such design not only disregards their unique characteristics but also may lead to information confusion, thus impairing the accurate understanding and processing of visual information. 
On the other hand, freezing visual encoder directly limits the ability to learn and represent visual information. 
Therefore, improving the visual perception requires rethinking the decoder design and optimizing the use of the visual encoder to better capture and process visual features.

As shown in Fig.~\ref{fig:motivation}(c), previous multimodal large language models (MLLMs) typically relied on CLIP as the visual encoder. However, research~\cite{tong2024eyes,xu2024llava-uhd} has revealed limitations in CLIP's ability to capture complex visual features. To address this, recent methods have incorporated self-supervised learning (SSL) pretrained models, such as DINOv2~\cite{oquab2024dinov2}, and fused their outputs with CLIP’s features to enhance the visual encoder's representation capacity. While this fusion approach improves feature representation, it also introduces significant computational overhead. The use of two visual encoders doubles the number of visual tokens, leading to a substantial increase in training costs, particularly when handling large-scale datasets and complex models.

Toward this end, we propose a new multimodal large language model \textbf{Arcana} that aims to enhance visual perception capabilities from both visual encoder and decoder. Specifically, we design a multimodal LoRA (MM-LoRA) to construct a multimodal decoder as show in Fig.~\ref{fig:motivation}(b). This decoder provides independent learning spaces for each modality, ensuring the decoupling of different modalities, avoiding information confusion, and preserving the uniqueness of each modality. Additionally, we propose a novel design, the Query Ladder Adapter (QLadder), as shown in Fig.~\ref{fig:motivation}(c). Unlike methods that significantly increase the number of visual tokens, our approach introduces only a small set of visual tokens (set to $64$, where $64 << N$). Despite the limited number of tokens, QLadder effectively enhances the model's ability to learn and represent visual information by progressively refining and integrating visual features through its \textit{"ladder"} structure. Notably, even with the introduction of only a small number of visual tokens, QLadder achieves performance comparable to DINOv2-based MOF~\cite{tong2024eyes} methods on the MMVP benchmark, which demands strong visual representations. Furthermore, our approach demonstrates performance improvements on traditional multimodal benchmarks, such as MMbench~\cite{liu2023mmbench} and TextVQA~\cite{textvqa}, highlighting its versatility and effectiveness across various tasks. 

Finally, we conducted an extensive series of experiments to thoroughly evaluate the performance and effectiveness of Arcana. These experiments were designed to assess various aspects, including the robustness of MM-LoRA and QLadder across different benchmarks, its ability to generalize in diverse scenarios, and its performance in comparison to state-of-the-art methods.


\vspace{-1.6em}
~\vspace{-1.6em}
\section{Related Work}
\vspace{-.6em}
\textbf{Multi-Modal Large Language Models.}
Fueled by the tremendous success of large language models (LLMs)~\cite{achiam2023gpt,touvron2023llama,jiang2023mistral}, there is growing interest in developing end-to-end multi-modal large language models (MLLMs)~\cite{dai2024instructblip,ye2023mplug,dong2024internlm}. These models aim to enhance the visual perceptual capabilities of LLMs by integrating additional modalities, allowing for unified handling of multi-modal tasks. Currently, there are three primary approaches to building Multi-Modal foundational models, each demonstrating strong potential for zero-shot generalization in the visual-language domain.

The first approach, exemplified by Flamingo~\cite{alayrac2022flamingo}, uses cross-attention to align visual models with large language models across modalities. The second approach, used by models like PaLM-E~\cite{driess2023palm}, directly integrates extracted visual features into a pre-trained PaLM~\cite{anil2023palm} model via a linear layer, achieving robust performance. This method is widely adopted by mainstream models such as LLaVA~\cite{llava}, CogVLM~\cite{wang2023cogvlm} and Internlm-Xcomposer~\cite{zhang2023xcomposer} but incurs high inference costs due to the lengthy visual tokens. To address this, the third approach, inspired by DETR~\cite{meng2021conditional,zhu2020deformable} and represented by BLIP-2~\cite{li2022blip}, employs a Q-former to effectively reduce the sequence length of visual features. Similar designs are used by mPLUG-OWL2~\cite{ye2023mplug}, and MiniGPT-4~\cite{zhu2023minigpt}. However, these methods~\cite{anil2023palm,bai2023qwen,chen2023shikra} couple visual and language modalities in the same space using language-guided decoders, overlooking the uniqueness of different modalities. This oversight may result in interference between modalities, potentially affecting performance. To this end, we employ MM-LoRA to implement a multimodal decoder, aiming to mitigate the impact of modality interference on the model.

\textbf{Improve visual perception for MLLMs.} Currently, MLLMs are the most popular approach in VL community~\cite{alayrac2022flamingo,li2022blip}, and enhancing their visual recognition capabilities has become a prominent research trend. Integrating visual features into large language models (LLMs) via a linear layer has become the mainstream approach~\cite{llava,wang2023cogvlm}. However, this approach often relies on frozen vision encoders to provide visual features, which limits the visual recognition capabilities of multimodal large language models (MLLMs). To address this issue, existing methods enhance visual recognition in two ways. The first method~\cite{luo2024feast,tong2024eyes,xu2024llava-uhd} introduces new high-resolution vision encoders, significantly improving visual recognition by increasing the number of visual tokens. For example, LLaVA-HR~\cite{luo2024feast} achieves this by incorporating ConvNeXt~\cite{liu2022convnet} to handle high-resolution images. However, these methods significantly increases the number of visual tokens. Therefore, we propose QLadder, which can significantly enhance the model's visual perception capability with the introduction of a small number of visual tokens. The second method~\cite{wang2023cogvlm,dong2024internlm,ye2023mplug} expands the learning space for visual tokens within the large language model to accelerate visual-language alignment, thereby enhancing visual perception. For instance, Internlm-Xcomposer2~\cite{dong2024internlm} introduces Partial-LoRA, adding a LoRA to visual tokens to strengthen their representation. However, experiments with MM-LoRA have shown that directly increasing the learning space for visual tokens in the decoder does not improve the model's performance.

~\vspace{-2.3em}
\section{Method}

\begin{figure}
	\centering
	\includegraphics[width=\linewidth]{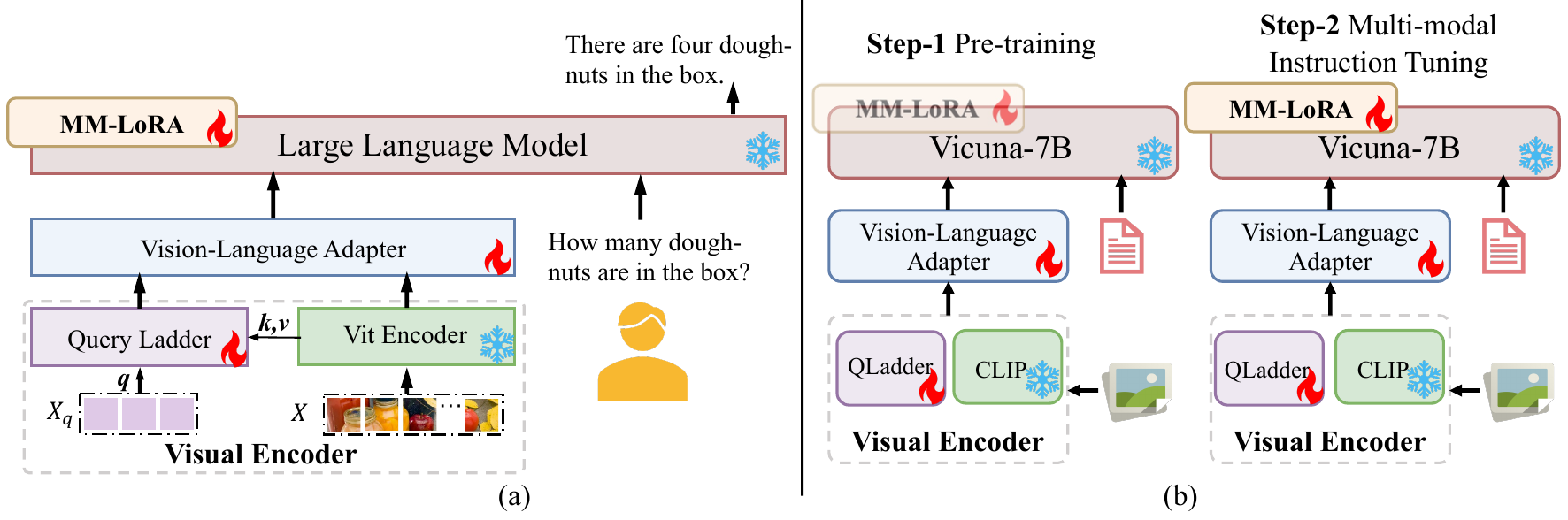}
	\vspace{-1.2em}
	\caption{\textbf{(a)} The architecture of the Arcana. \textbf{(b)} The training pipeline of  Arcana. MM-LoRA is optional during the pre-training phase.}
	\label{fig:arch_train}
 \vspace{-2.3em}
\end{figure}
\vspace{-.9em}
\subsection{Overview}
We propose a new model, named Arcana as shown in Fig~\ref{fig:arch_train}, designed to enhance visual perception in multimodal language models. Like most existing models~\cite{llava,chen2023shikra}, it includes a visual encoder, a vision-language adapter, and a large language model.
The key difference is that we use MM-LoRA to implement a multimodal decoder. Unlike traditional fine-tuning where visual and language modalities share LoRA parameters, MM-Lora assigns different LoRA parameters to each modality. Additionally, we introduce QLadder in the visual encoder, which significantly enhances the model's ability to learn and represent visual information with the introduction of a small number of visual tokens. We first briefly introduce Arcana’s architecture in Section~\ref{sec: arch}. Additionally, in Section~\ref{sec: slora}, we detail MM-LoRA, which aims to decouple the learning spaces of different modalities to achieve a multimodal decoder. Lastly, we introduce the training paradigm of Arcana in Section~\ref{sec: train}.

\subsection{Architecture}\label{sec: arch}
\vspace{-.4em}
Our approach Arcana 
(illustrated in Fig.~\ref{fig:arch_train}(a) consists of three main components: 
{visual encoder}, 
{vision-language adapter} and 
{large language model}. 
Each component is described in the following.

\begin{figure}
	\centering
	\includegraphics[width=0.95\linewidth]{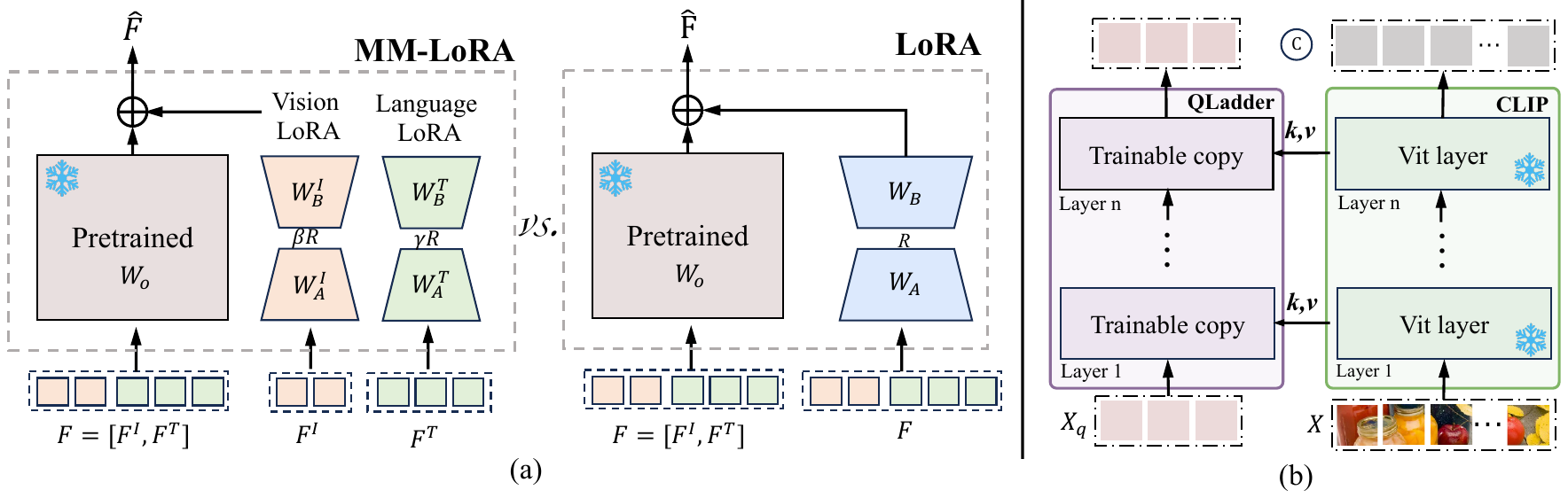}
	\vspace{-0.7em}
	\caption{\textbf{(a)} The farmework of MM-LoRA \emph{vs.} LoRA. MM-LoRA introduces two new hyperparameters, $\beta$ and $\gamma$, to control the ranks of the visual and language LoRAs, respectively. Notably, we set $\beta + \gamma = 1$ to ensure that MM-LoRA has the same number of parameters as LoRA. \textbf{(b)} The architecture of the visual encoder includes the QLadder adapter and CLIP. The QLadder adapter consists of cross-attention and FFN layers, with weights initialized from those of CLIP. }
	\label{fig:slora_qladder}
 \vspace{-1.5em}
\end{figure}

\textbf{Visual Encoder.} The primary objective is to extract visual features from the image. 
The encoder learned with language-supervision,
e.g., the CLIP~\cite{radford2021learning} visual model,
is widely adopted. 
The CLIP encoder is often fixed during fine-tuning,
e.g., in LLaVA,
for keeping the representation capability
of the original CLIP encoder.
We propose to
improve the visual encoder
through a query ladder adaptor (QLadder)
from the fine-tuning data
that may contain new visual semantics. 
The structure is shown in Fig.~\ref{fig:slora_qladder}(b). This adapter enhances the visual feature representation of the visual encoder by adding a small number of query visual tokens while retaining the pretrained image encoder. It improves Arcana's visual perception capability.


We extract visual features $\mathbf{F}_c\in \mathbb{R}^{N_I\times C_v}$
through the CLIP encoder, where $C_v$ represents the channel of visual feature, $N_I$ indicates the number of image patch. 
A set of learnable vectors $\mathbf{x}_q$ is fed into QLadder to 
acquire additional visual features $\mathbf{F}_q\in \mathbb{R}^{N_q\times C_v}$, where $N_q << N_I$. 
The two kinds of visual features are concatenated: $\mathbf{F}_v = \operatorname{concat}(\mathbf{F}_c, \mathbf{F}_q)$. As shown in Fig~\ref{fig:arch_train}(b), QLadder comprises multiple layers composed of cross-attention and feed-forward networks (FFNs). 


\textbf{Vision-Language Adapter.} 
We map the output of visual encoder to the same space as the language embedding space
through an Vision-Language adapter.
The adapter $\operatorname{g}$ consists of two MLP layers. 
The output visual features are denoted as 
$\mathbf{F}^I = \operatorname{g}(\mathbf{F}_v)$.

\textbf{Large Language Model.} For multimodal tasks~\cite{goyal2017vqav2,hudson2019gqa}, leveraging pre-trained large language models (LLMs) can provide valuable linguistic priors. Through multimodal instruction tuning, LLMs learn to comprehend visual features within images, enabling comprehensive understanding and processing of multimodal data. Typically, this process is accomplished through full fine-tuning or LoRA~\cite{hu2021lora}. However, these methods overlook the unique characteristics of modalities, leading to modality confusion. This not only damages MLLMs' accurate understanding and processing of visual information but also affects natural language understanding. Therefore, a multimodal decoder that provides separate learning spaces for each modality is a better choice for MLLMs.

\subsection{Multimodal LoRA}\label{sec: slora}
To implement a multimodal decoder based on a large language model, we propose a multimodal LoRA. This approach projects visual and language features into separate semantic spaces to decouple their representations, thereby avoiding modality interference. This allows Arcana to retain the unique characteristics of each modality, enhancing its visual perception without compromising natural language understanding. Next, we detail the MM-LoRA process.

MM-LoRA, as illustrated in Fig.~\ref{fig:slora_qladder}, consists of visual LoRA and language LoRA. In comparison to LoRA, we introduces two parameters, $\beta$ and $\gamma$, to control the rank size of ($R$) visual LoRA and language LoRA. It's worth noting that $\beta + \gamma = 1$ to ensure that no additional parameters are introduced compared to LoRA. Specifically, given a sequence of visual-language features $F \in \mathbb{R}^{(N_v+N_t) \times C}$ and a multimodal mask $M\in \{0,1\}^{(N_v+N_t)}$, where $C$ represents the  hidden dimension in LLMs, $N_v$ and $N_t$ indicates the number of visual and language tokens, respectively. We define a modality separation function $\Theta$ to separate the tokens of different modalities within $F$.
\begin{equation}
    \Theta(F, M, m) = F\odot (M==m)\,,
\end{equation}
where $m\in \{0,1\}$ is used to select between visual tokens (\(m=0\)) and language tokens (\(m=1\)). Therefore, based on multimodal mask $M$ , we can obtain $F^I$ and $F^T$.
\begin{equation}
    F^I = \Theta(F, M, 0) \qquad F^T = \Theta(F, M, 1)
\end{equation}
Then, $F^I$ and $F^T$ are separately inputted into the visual part and language part of MM-LoRA. In Visual LoRA, the weights are denoted as $W_A^I\in \mathbb{R}^{C \times \beta R}$ and $W_B^I\in \mathbb{R}^{\beta R \times C}$, while in Language LoRA, the weights are denoted as $W_A^T\in \mathbb{R}^{C \times \gamma R}$ and $W_B^T\in \mathbb{R}^{\gamma R \times C}$. 

Similarly to LoRA~\cite{hu2021lora}, $F$ is inserted into the LLM layer to obtain $\hat{F}$. Finally, the output results of MM-LoRA are added to the output of LLM according to the mask $M_I$.

\vspace{-4mm}
\begin{equation}
\begin{split}
\hat{F} &= W_o\times F \\
\Theta(\hat{F}, M, 0) += W_B^I\times W_A^I\times F^I &\qquad \Theta(\hat{F}, M, 1) += W_B^T\times W_A^T\times F^T \\
\end{split}
\end{equation}

In Arcana, MM-LoRA is applied to all linear layers of the large language model, thereby achieving an optimal multimodal decoder.

\subsection{Training Paradigm}\label{sec: train}
Following prior work~\cite{llava,wang2023cogvlm}, we adopt a two-stage approach involving pretraining and multimodal instruction fine-tuning to train Arcana, as illustrated in Fig.~\ref{fig:arch_train}(b). The purpose of the pretraining stage is to align the visual encoder with the language model, while multimodal instruction fine-tuning aims to adapt the model better to specific tasks through fine-tuning. We found that freezing the visual encoder limits the MLLM's ability to capture complex visual information, such as scene text and visual knowledge. To address this issue, we introduce Qladder and enable it to be trained in both the pretraining and instruction fine-tuning stages. This strategy allows the model to more effectively capture both low-level and high-level semantic visual information. Additionally, we introduce MM-LoRA fine-tuning as an alternative to full fine-tuning and LoRA fine-tuning, enabling a multimodal decoder that minimizes modality interference. Specifically, in the pretraining stage, we train Qladder and the vision-language adapter, while in the instruction fine-tuning stage, we train Qladder, the vision-language adapter, and MM-LoRA. Furthermore, to ensure the linguistic capabilities of Arcana, we employ joint training, adjusting the entire model during instruction fine-tuning, integrating textual and multimodal instructions.



~\vspace{-1.1em}
\section{Experiments}\label{sec:experiments}
\vspace{-.4em}
\subsection{Implementation Details}
\textbf{Model.} In the visual encoder, we utilize the CLIP-L~\cite{radford2021learning} model with an input resolution of 336 and a patch size of $14\times 14$. Furthermore, the QLadder adapter adopts the same structure as CLIP-L,  replacing self-attention with cross-attention. Notably, QLadder utilizes pre-trained CLIP weights as its initial weights. For the LLMs, we employ the pre-trained Vicuna-7B~\cite{chiang2023vicuna} model. The Vision-Language adapter comprises two layer MLP. MM-LoRA, used for fully supervised multimodal instruction tuning, consists of a visual LoRA with a rank of $\beta \times R $ and a language LoRA with a rank of $\gamma \times R$. 
\begin{table*}[!t]
\centering
\setlength\tabcolsep{4pt}
\renewcommand\arraystretch{1.3}
\setlength{\tabcolsep}{1.5mm}{
\caption{Performance on six General Visual Question Answering benchmarks. Specialist models, indicated in \textcolor{lightgray}{gray}, are fine-tuned on each individual dataset. The \textcolor{red}{red} and \textcolor{blue}{blue} colors respectively represent the optimal and suboptimal results on each benchmark. $*$ indicates that MM-LoRA is trained during the pretrain stage. }
\vspace{+.5em}
\label{tab:vqa}
\resizebox{\textwidth}{!}{%
\begin{tabular}{c|l|c|ccc|ccc}
\hline
\multirow{2}{*}{Type} & \multirow{2}{*}{Model} & \multirow{2}{*}{LLM} & \multicolumn{3}{c|}{In-domain VQA Tasks} & \multicolumn{3}{c}{Zero-shot VQA Tasks} \\ \cline{4-9}
 &  &  & VQAv2 & OKVQA & GQA & TextVQA & ScienceQA & Ai2d \\ \hline
\multirow{10}{*}{Generalists} & BLIP2~\cite{li2022blip} & Flan-T5 & 65.0 & 45.9 & 41.0 & 42.5 & 61.0 & - \\
 & InstructBLIP~\cite{dai2024instructblip} & Vicuna (7B) & - & - & 49.2 & 50.1 & 60.5 & 40.6 \\
 & InstructBLIP~\cite{dai2024instructblip} & Vicuna (13B) & - & - & 49.5 & 50.7 & 63.1 & - \\
 & Shikra~\cite{chen2023shikra} & Vicuna (7B) & 77.4 & 47.2 & - & - & - & - \\
 & IDEFICS-Instruct~\cite{laurenccon2024obelics} & LLaMA (65B) & 37.4 & 36.9 & - & 28.3 & 61.8 & 54.8 \\  
 & LLaVA-v1.5~\cite{liu2023llava1.5} & Vicuna (7B) & \multicolumn{1}{c}{78.5} & \multicolumn{1}{c}{-} & {\textcolor{red}{\textbf{62.0}}} & \multicolumn{1}{c}{58.2} & \multicolumn{1}{c}{66.8} & 55.5 \\ 
 & Qwen-VL-Chat~\cite{bai2023qwen} & Qwen (7B) & \multicolumn{1}{c}{78.2} & \multicolumn{1}{c}{56.6} & 57.5 & \multicolumn{1}{c}{{\textcolor{red}{\textbf{61.5}}}} & \multicolumn{1}{c}{68.2} & - \\  
 & mPLUG-Owl2~\cite{ye2023mplug} & LLaMA (7B) & \multicolumn{1}{c}{\textcolor{blue}{\textbf{79.4}}} & \multicolumn{1}{c}{57.7} & 56.1 & \multicolumn{1}{c}{58.2} & \multicolumn{1}{c}{68.7} & 55.7 \\ \cline{2-9} 
 & Arcana & Vicuna (7B) & \multicolumn{1}{c}{79.2} & \multicolumn{1}{c}{\textcolor{blue}{\textbf{57.9}}} & 61.6 & \multicolumn{1}{c}{{\textcolor{blue}{\textbf{59.5}}}} & \multicolumn{1}{c}{{\textcolor{red}{\textbf{71.2}}}} & \textcolor{blue}{\textbf{56.8}} \\
 & Arcana$^*$ & Vicuna (7B) & \multicolumn{1}{c}{\textcolor{red}{\textbf{79.5}}} & \multicolumn{1}{c}{\textcolor{red}{\textbf{58.9}}} & {\textcolor{blue}{\textbf{61.8}}} & \multicolumn{1}{c}{58.7} & \multicolumn{1}{c}{{\textcolor{blue}{69.5}}} & \textcolor{red}{\textbf{56.9}} \\ \hline
\multirow{2}{*}{\textcolor{lightgray}{Specialists}} 
 & \textcolor{lightgray}{GIT2~\cite{wang2022git}} & \textcolor{lightgray}{-} & \multicolumn{1}{c}{\textcolor{lightgray}{81.7}} & \multicolumn{1}{c}{\textcolor{lightgray}{-}} & \textcolor{lightgray}{-} & \multicolumn{1}{c}{\textcolor{lightgray}{59.8}} & \multicolumn{1}{c}{\textcolor{lightgray}{-}} & \textcolor{lightgray}{-} \\ 
 & \textcolor{lightgray}{PaLI-17B~\cite{chen2022pali}} & \textcolor{lightgray}{-} & \multicolumn{1}{c}{\textcolor{lightgray}{84.3}} & \multicolumn{1}{c}{\textcolor{lightgray}{64.5}} & \textcolor{lightgray}{-} & \multicolumn{1}{c}{\textcolor{lightgray}{58.8}} & \multicolumn{1}{c}{\textcolor{lightgray}{-}} & \textcolor{lightgray}{-} \\ \hline
\end{tabular}
}}
\vspace{-1.em}
\end{table*}


\textbf{Data Sets.} During pre-training, we used approximately 1.2M image-text pairs from ShareGPT4V~\cite{chen2023sharegpt4v}. In the multimodal instruction tuning stage, we utilize six types of supervised data totaling 934k, namely: (1) text-only instruction data (ShareGPT~\cite{sharegpt}); (2) vision question-answering data (VQAv2~\cite{goyal2017vqav2}, GQA~\cite{hudson2019gqa}, A-OKVQA~\cite{schwenk2022okvqa}, OK-VQA~\cite{marino2019ok}); (3) OCR QA (OCRVQA~\cite{mishra2019ocr}, TextCaps~\cite{sidorov2020textcaps}); (4) Region-aware QA (RefCOCO~\cite{kazemzadeh2014referitgame,mao2016generation}, VG~\cite{krishna2017visual}); (5) multi-modal instruction data (LLaVA-instruct~\cite{llava}); and (6) image captions (VG-COCO~\cite{hao2024fullanno}, shareGPT4V~\cite{chen2023sharegpt4v}). In the Ablation study, we only use the multimodal instruction data from LLaVA-v1.5.

\textbf{Training Setting.} During the pretraining step, we use language modeling loss with a batch size of $256$ for $1$ epoch. The learning rates are set to $1e-3$ for the vision-language adapter and $2e-5$ for Qladder. In the multimodal instruction tuning step, we integrated MM-LoRA into the LLM to create a multimodal decoder, thus preventing information interference between modalities. We set the learning rate for MM-LoRA to $1e-4$, and for both Qladder and the vision-language adapter, to $2e-5$. MM-LoRA is configured with a default rank $R$ of 256, $\beta$ set to 0.25, and $\gamma$ set to 0.75. All experiments are conducted on 8 NVIDIA A100 GPUs.

\subsection{Main Results}

\begin{table*}[!t]
\centering
\setlength\tabcolsep{4pt}
\renewcommand\arraystretch{1.5}
\setlength{\tabcolsep}{2mm}{
\caption{Performance on five Large Vision-Language Models (LVLM) benchmarks.The \textcolor{red}{red} and \textcolor{blue}{blue} colors respectively represent the optimal and suboptimal results on each benchmark. $*$ indicates that MM LoRA is trained during the pretrain stage.}
\vspace{+.5em}
\label{tab:lvlm}
\resizebox{\textwidth}{!}{%
\begin{tabular}{l|c|c|cccccc}
\hline
Method & Vision Encoder & Language Model &MME & MMBench & MM-Vet & SEED-Bench & LLaVA$^W$ & POPE \\ \hline
BLIP-2~\cite{li2022blip} & ViT-g (1.3B) & Vicuna (7B) & 1293.84 & - & 22.4 & 46.4 & 38.1 & 85.3 \\
MiniGPT-4~\cite{zhu2023minigpt} & ViT-g (1.3B) & Vicuna (7B) & 581.67  & 23.0 & 22.1 & 42.8 & 45.1 & - \\
LLaVA~\cite{llava} & ViT-L (0.3B) & Vicuna (7B) & 502.82 & 36.2 & 28.1 & 33.5 & 63.0 & 80.2 \\
mPLUG-Owl~\cite{ye2023mplug} & ViT-L (0.3B) & LLaMA (7B) & 967.34 & 46.6 & - & 34.0 & - & - \\
InstructBLIP~\cite{dai2024instructblip} & ViT-g (1.3B) & Vicuna (7B) & 1212.82 & 36.0 & 26.2 & 53.4 & 60.9 & 78.9 \\
LLaMA-Adapter-v2~\cite{gao2023llamaadapter2} & ViT-L (0.3B) & LLaMA (7B) & 1328.40 & 39.5 & 31.4 & 32.7 & - & - \\
Otter~\cite{li2023otter} & ViT-L (0.3B) & LLaMA (7B) & 1292.26  & 48.3 & 24.6 & 32.9 & - & - \\
Qwen-VL-Chat~\cite{bai2023qwen} & ViT-G (1.9B) & Qwen (7B) & 1487.58 & 60.6 & - & 58.2 & - & - \\
LLaVA-v1.5~\cite{liu2023llava1.5} & ViT-L (0.3B) & Vicuna (7B) &\textcolor{blue}{\textbf{1510.70}} & 64.3 & 30.5 & 58.6 & 63.4 & 85.9 \\
mPLUG-Owl2~\cite{ye2023mplug2} & ViT-L (0.3B) & LLaMA (7B) & 1450.19 & 64.5 & \textcolor{red}{\textbf{36.2}} & 57.8 & - & 86.2 \\ \hline
\textbf{Arcana} & ViT-L (0.3B) & Vicuna (7B) & 1476.48 & \textcolor{blue}{\textbf{66.9}} & \textcolor{blue}{\textbf{34.8}} & \textcolor{blue}{\textbf{62.6}} & \textcolor{blue}{\textbf{67.3}} & \textcolor{blue}{\textbf{86.5}} \\
\textbf{Arcana$^*$} & ViT-L (0.3B) & Vicuna (7B) &\textcolor{red}{\textbf{1520.93}} & \textcolor{red}{\textbf{67.4}} & 34.4 & \textcolor{red}{\textbf{63.2}} & \textcolor{red}{\textbf{72.7}} & \textcolor{red}{\textbf{87.1}} \\ \hline
\end{tabular}
}}
\vspace{-1em}
\end{table*}
\textbf{General Visual Question Answering Benchmarks.} In Table~\ref{tab:vqa}, we compare with both SOTA MLLMs model on six General VQA benchmarks, including VQAv2~\cite{goyal2017vqav2}, OKVQA~\cite{schwenk2022okvqa}, GQA~\cite{hudson2019gqa}, TextVQA~\cite{textvqa}, ScienceQA~\cite{scienceQA} and Ai2d~\cite{ai2d}. We found that Arcana achieved competitive results on six VQA benchmarks. Notably, it achieved accuracies of 57.9 on OKVQA, 71.2 on ScienceQA, and 56.8 on Ai2d , surpassing most recently proposed MLLMs methods. Additionally, Arcana$^*$ with MM-LoRA used during the pre-training stage achieved better performance, indicating the importance of preserving the uniqueness of different modalities during pre-training. The superior performance on zero-shot VQA tasks particularly highlights strong generalization ability and potential across different domains of our model.

\textbf{Large Vision-Language Model Benchmarks.} Table~\ref{tab:lvlm} presents our comparative results on five different LVLM benchmarks: MMBench~\cite{liu2023mmbench}, MM-Vet~\cite{yu2023mmvet}, SEED-Bench~\cite{li2023seed-bench}, LLava$^W$~\cite{llava}, and POPE~\cite{li2023pope}. It is evident that Arcana achieves highly competitive performance across these benchmarks. Compared to mPLUG-OWL2~\cite{ye2023mplug2}, Arcana scores 2.4 and 4.8 points higher on MMBench and SEED-Bench, respectively. Additionally, Arcana achieves a score of 86.5 on the hallucination evaluation dataset POPE, indicating significant advancements in visual recognition capabilities. These impressive results not only demonstrate its strong reasoning and multi-task generalization abilities but also clearly show that Arcana significantly outperforms others in these areas. Notably, we achieved this using a 0.3B visual encoder, with MM-LoRA and QLadder significantly enhancing the model's visual perception and generalization.

\begin{wrapfigure}{r}{7.5cm}
\vspace{-0.7cm}
\centering
\begin{minipage}{1.0\linewidth}
\begin{table}[H]
\centering
\setlength\tabcolsep{4pt}
\renewcommand\arraystretch{1.6}
\setlength{\tabcolsep}{4.2mm}{
\caption{\textbf{Performance on language benchmarks of our model} compared to LLaMA-2 0-shot for BBH, AGIEval, ARC.}
\label{tab:NLP}
\resizebox{\textwidth}{!}{%
        \begin{tabular}{l|c|c|c|c}
        \hline
            Method & BBH & AGIEval & ARC-c & ARC-e  \\ \cline{1-5}
            LLaMA-2~\cite{touvron2023llama2}  & 38.2 & 21.8 & 40.3 & 56.1 \\
            WizardLM~\cite{xu2023wizardlm} & 34.7 & 23.2 & 47.5 & 59.6 \\
            LLaMA-2-Chat~\cite{touvron2023llama2}  & 35.6 & 28.5 & 54.9 & 71.6 \\
            Vicuna-v1.5~\cite{chiang2023vicuna} & 41.2 & 21.2 & 56.6 & 72.8 \\ \cline{1-5}
            \textbf{Arcana} & \textbf{42.1} & \textbf{29.3} & \textbf{61.4} & \textbf{78.3} \\ \hline
    \end{tabular}
    }
    }  
\end{table}
\end{minipage}
\end{wrapfigure}

\textbf{Natural Language Understanding.} Although MLLMs excel in various multimodal downstream tasks, existing work~\cite{llava, dong2024internlm} often overlooks their natural language understanding capabilities. To address this, we also evaluated our model's language understanding performance on BIG-Bench Hard (BBH)~\cite{suzgun2023bbh}, AGIEval~\cite{zhong2023agieval}, and ARC~\cite{clark2018ARC}, as shown in Table~\ref{tab:NLP}. Compared to LLaMA-like~\cite{touvron2023llama} language models, Arcana achieved competitive results across multiple benchmarks. This demonstrates that our model not only performs well in multimodal tasks but also excels in language understanding, further highlighting the superiority of our approach.

\vspace{-.6em}
\subsection{Ablation Study}
\vspace{-.2em}
To validate the effectiveness of QLadder and MM-LoRA, we designed a series of experiments. Additionally, to ensure fairness, we used only LLaVA-v1.5~\cite{liu2023llava1.5} data for these experiments.\\

\textbf{Multimodal LoRA (MM-LoRA).} To validate the effectiveness of the multimodal decoder, we compared the performance of MM-LoRA and LoRA. Additionally, to investigate the importance of visual tokens and language tokens in the multimodal instruction tuning process within the decoder, we compared different ratios of $\beta$ and $\gamma$ parameters. In all experiments, the RANK of MM-LoRA and LoRA was set to 256. The results are shown in Table~\ref{tab:mmlora}. It clearly indicate that MM-LoRA achieves optimal performance when $\beta=0.25$ and $\gamma=0.75$. When $\beta$ is set to 1, performance significantly drops, indicating that aligning language distribution using only visual tokens is challenging for\\

\begin{minipage}{\textwidth}
\vspace{-1.2em}
\begin{minipage}{0.6\textwidth}\scriptsize
\makeatletter\def\@captype{table}\makeatother
\setlength\tabcolsep{4pt}
\renewcommand\arraystretch{1.4}
\centering
\setlength{\tabcolsep}{1.4mm}{
\caption{Ablation of $\beta$ and $\gamma$ in MM-LoRA. The default rank is set to 256, while $\beta$ and $\gamma$ are used to control the rank values in visual and language LoRA components, respectively.}
\vspace{+.8em}
\label{tab:mmlora}
\resizebox{\textwidth}{!}{%
\begin{tabular}{c|cc|cccc}
\hline
\multirow{2}{*}{Method} & \multicolumn{2}{c|}{RANK} & \multirow{2}{*}{TextVQA} & \multirow{2}{*}{ScienceQA} & \multirow{2}{*}{MMBench} & \multirow{2}{*}{MME} \\ \cline{2-3}
 & $\beta$ & $\gamma$ &  &  &  &  \\ \hline
LoRA & - & - & 58.1 & 69.1 & 63.8 & 1460 \\ \hline
\multirow{5}{*}{MMLoRA} & 1 & 0 & 51.2$_{\textcolor{green}{(-6.9)}}$ & 65.8$_{\textcolor{green}{(-3.3)}}$ & 56.4$_{\textcolor{green}{(-7.4)}}$ & 1356$_{\textcolor{green}{(-104)}}$ \\
 & 0.75 & 0.25 & \textbf{58.7}$_{\textcolor{red}{(+0.6)}}$ & 68.6$_{\textcolor{green}{(-0.5)}}$ & 63.3$_{\textcolor{green}{(-0.5)}}$ & 1465$_{\textcolor{red}{(+5.0)}}$ \\
 & 0.5 & 0.5 & 58.5$_{\textcolor{red}{(+0.4)}}$ & 70.1$_{\textcolor{red}{(+1.0)}}$ & 64.4$_{\textcolor{red}{(+0.6)}}$ & 1483$_{\textcolor{red}{(+23)}}$ \\
 & 0.25 & 0.75 & \textbf{58.7}$_{\textcolor{red}{(+0.6)}}$ & \textbf{71.2}$_{\textcolor{red}{(+2.1)}}$ & 64.8$_{\textcolor{red}{(+1.0)}}$ & \textbf{1500}$_{\textcolor{red}{(+40)}}$ \\
 & 0 & 1 & 57.9$_{\textcolor{green}{(-0.2)}}$ & 70.1$_{\textcolor{red}{(+1.0)}}$ & \textbf{65.4}$_{\textcolor{red}{(+1.6)}}$ & 1480$_{\textcolor{red}{(+20)}}$ \\ \hline
\end{tabular}
}}
\end{minipage}\hspace{0.2em}
\begin{minipage}{0.4\textwidth}
\centering
\setlength\tabcolsep{4pt}
\makeatletter\def\@captype{table}\makeatother
\renewcommand\arraystretch{1.8}
\setlength{\tabcolsep}{1.4mm}{
\caption{Ablation of query number in QLadder. N$_q$ represents the number of learnable query.}
\vspace{+.8em}
\label{tab:num_query}
\resizebox{\textwidth}{!}{%
\begin{tabular}{c|c|ccc}
\hline
Method & N$_q$ & ScienceQA & MMBench & MME \\ \hline
baseline & - & 69.1 & 63.8 & 1460 \\ \hline
\multirow{4}{*}{+QLadder} & 16 & 70.4$_{\textcolor{red}{(+1.3)}}$ & 63.9$_{\textcolor{red}{(+0.1)}}$ & 1481$_{\textcolor{red}{(+21)}}$ \\
 & 32 & 70.6$_{\textcolor{red}{(+1.5)}}$ & 64.6$_{\textcolor{red}{(+0.8)}}$ & 1493$_{\textcolor{red}{(+33)}}$ \\
 & 64 & \textbf{71.2}$_{\textcolor{red}{(+2.1)}}$ & \textbf{64.8}$_{\textcolor{red}{(+1.0)}}$ & \textbf{1500}$_{\textcolor{red}{(+40)}}$ \\
 & 128 & 69.7$_{\textcolor{red}{(+0.6)}}$ & 64.2$_{\textcolor{red}{(+0.4)}}$ & 1473$_{\textcolor{red}{(+13)}}$ \\ \hline
\end{tabular}
}}
\end{minipage}

\end{minipage}

MLLMs. However, introducing $\gamma$ greatly improves performance, demonstrating that learning both vision and language simultaneously accelerates modality alignment. When $\gamma$ is set to 1, there is a slight performance decline, but MM-LoRA still matches LoRA's performance, suggesting that visual token learning is less critical than language token learning in LLMs. This indicates that during the instruction tuning phase of MLLM training, more emphasis should be placed on learning language tokens. Furthermore, when both $\beta$ and $\gamma$ are set to 0.5, the performance of MM-LoRA significantly outperforms LoRA. This intuitively demonstrates that the multimodal decoder can avoid interference between modalities by separating them, thus significantly enhancing the performance of MLLMs. \\

\begin{figure*}
	\centering
	\includegraphics[width=\linewidth]{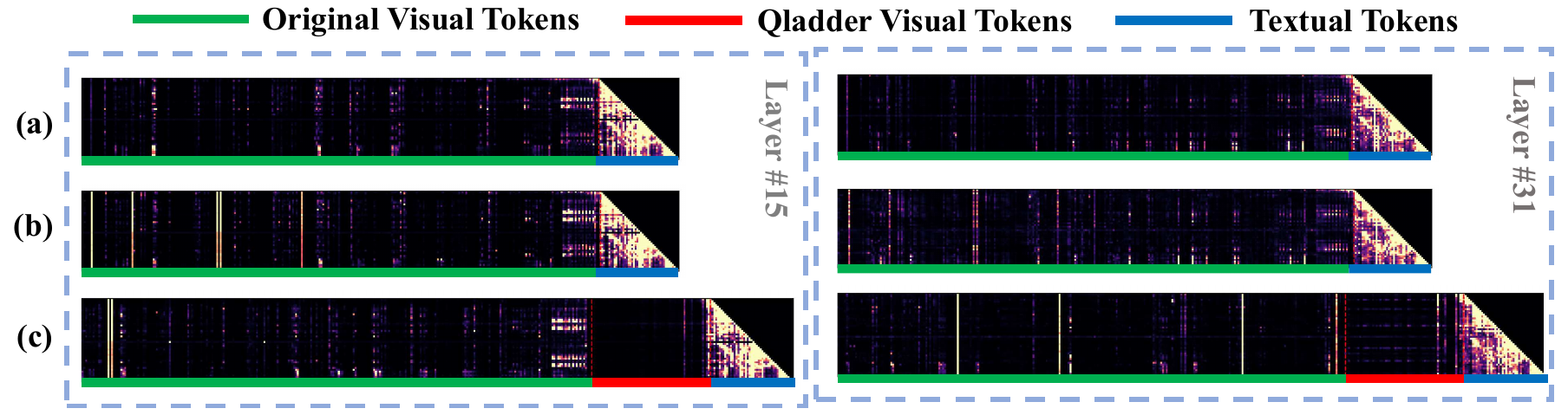}
	\vspace{-1.2em}
	\caption{\textbf{Visualization of attention maps}. We compare the attention maps in different layer of LLM between different composition, include \textbf{(a)} Baseline, \textbf{(b)}Baseline+MM-LoRA, and \textbf{(c)} Baseline+MM-LoRA+QLadder. Higher brightness indicates higher attention values, with the x-axis representing all tokens, and the y-axis containing only the generated text tokens.}
	\label{fig:attention}
 \vspace{-1.2em}
\end{figure*}

\vspace{-.4em}
\textbf{QLadder in Vision Encoder.} To validate the effectiveness of QLadder and determine the optimal number of queries, we conducted experiments with QLadder. The results, shown in Table~\ref{tab:num_query}, indicate that the inclusion of QLadder significantly enhances our model's performance. This demonstrates that even with a slight increase in visual tokens, without introducing a new visual encoder, the model's visual recognition capabilities can be improved.  As the number of queries increased, our model's performance gradually improved, reaching its best performance with 64 queries. However, further increasing the number of queries led to a performance decline, indicating that too many queries can negatively impact the model's performance.
To explicitly demonstrate the computational costs and efficiency of MLLMs with and without QLadder, we tested the memory usage and inference speed under both setting. As shown in Table~\ref{tab:computation}, even with QLadder, MLLMs only increase memory usage by 0.582G, and the inference speed decreases by just 0.11 tokens/s. 
This shows that the additional computational costs and efficiency impacts of QLadder are minimal and acceptable given the improvements it brings.

\textbf{QLadder tuning v.s. Visual Encoder tuning.}
To explore the impact of fine-tuning QLadder versus directly fine-tuning the Visual Encoder, we conducted comparative experiments to evaluate the effects of tuning the vision encoder, freezing the vision encoder, and adding Q-Ladder. The results are shown in Table~\ref{tab:visual_tuning}.
Tuning the Vision Encoder often leads to the loss of pre-trained knowledge and does not significantly enhance MLLM's performance. In some benchmark tests, it may even have a negative impact.
Freezing the Vision Encoder preserves pre-trained knowledge but lacks further optimization potential.
Adding Q-Ladder significantly improves MLLM's performance by enhancing visual feature representation with a small number of additional visual tokens, while retaining pre-trained knowledge.
These results demonstrate that Q-Ladder effectively strengthens visual feature representation and avoids the negative effects associated with tuning the vision encoder.

\begin{table*}[!t]
\centering
\setlength\tabcolsep{4pt}
\renewcommand\arraystretch{1.6}
\setlength{\tabcolsep}{4.5mm}{
\caption{Comparision with QLadder and additional Visual Encoder. To explore the performance in visual grounding ability, we selected MMVP, POPE, MMBench, and TextVQA for experiments. The data used in the experiments is consistent with that of LLaVA-v1.5.}
\vspace{+.5em}
\label{tab:additional_visual_encoder}
\resizebox{\textwidth}{!}{%
\begin{tabular}{l|cc|cccc}
\hline
Method & Size & add visual tokens & MMVP & POPE & MMBench & TextVQA \\ \hline
LLaVA-v1.5 & 7B & -  & 24.0 & 85.9 & 64.3 & 58.2 \\
LLaVA-v1.5 + MOF~\cite{tong2024eyes} & 7B & 256  & 27.1$_{\textcolor{red}{(+3.1)}}$ & 86.2$_{\textcolor{red}{(+0.3)}}$ & 60.1$_{\textcolor{green}{(-4.2)}}$ & 56.5$_{\textcolor{green}{(-1.7)}}$ \\
LLaVA-v1.5 + QLadder & 7B & 64  & 27.6$_{\textcolor{red}{(+3.6)}}$ & 86.5$_{\textcolor{red}{(+0.6)}}$ & 66.3$_{\textcolor{red}{(+2.0)}}$ & 58.8$_{\textcolor{red}{(+0.6)}}$ \\ \hline
 \textcolor{gray}{LLaVA-v1.5} &  \textcolor{gray}{13B} &  \textcolor{gray}{-}  &  \textcolor{gray}{24.7} &  \textcolor{gray}{85.9} &  \textcolor{gray}{67.7} &  \textcolor{gray}{61.3} \\
 \textcolor{gray}{LLaVA-v1.5 + MOF~\cite{tong2024eyes}} &  \textcolor{gray}{13B} &  \textcolor{gray}{256} &  \textcolor{gray}{28.0$_{\textcolor{red}{(+3.3)}}$} & \textcolor{gray}{86.3$_{\textcolor{red}{(+0.4)}}$} &  \textcolor{gray}{61.6$_{\textcolor{green}{(-6.1)}}$} &  \textcolor{gray}{55.3$_{\textcolor{green}{(-6.0)}}$} \\
 \textcolor{gray}{LLaVA-v1.5 + MOF~\cite{tong2024eyes}} &  \textcolor{gray}{13B} &  \textcolor{gray}{576} &  \textcolor{gray}{31.2$_{\textcolor{red}{(+6.5)}}$} &  \textcolor{gray}{86.7$_{\textcolor{red}{(+0.8)}}$} &  \textcolor{gray}{65.4$_{\textcolor{green}{(-2.3)}}$} &  \textcolor{gray}{58.7$_{\textcolor{green}{(-2.6)}}$} \\

\hline
\end{tabular}
}}
\vspace{-1em}
\end{table*}

\begin{minipage}{\textwidth}
\vspace{-1.2em}
\begin{minipage}{0.5\textwidth}\scriptsize
\makeatletter\def\@captype{table}\makeatother
\setlength\tabcolsep{4pt}
\renewcommand\arraystretch{1.4}
\centering
\setlength{\tabcolsep}{1.4mm}{
\caption{Comparing different tuning strategies for visual encoders.}
\vspace{+.8em}
\label{tab:visual_tuning}
\resizebox{\textwidth}{!}{%
\begin{tabular}{c|c|ccc}
\hline
Method & Vision Encoder & TextVQA & MMBench & MM-Vet \\ \hline
baseline & freezing & 58.1 & 64.1 & 31.5\\ 
baseline & tuning & 57.7 & 64.3 & 31.1\\ 
baseline & add QLadder & \textbf{58.8} & \textbf{66.3} & \textbf{33.7}\\ \hline
\end{tabular}
}}
\end{minipage}\hspace{0.2em}
\begin{minipage}{0.5\textwidth}
\centering
\setlength\tabcolsep{4pt}
\makeatletter\def\@captype{table}\makeatother
\renewcommand\arraystretch{1.8}
\setlength{\tabcolsep}{1.4mm}{
\caption{Comparison of computational load and resource utilization during inference.}
\vspace{+.8em}
\label{tab:computation}
\resizebox{\textwidth}{!}{%
\begin{tabular}{c|c|c}
\hline
Setting & Memory used & Inference Speed (token/s) \\ \hline
Arcana (w/o QLadder) & 15.243GB & 22.58 \\
Arcana (w QLadder) & 15.825GB & 22.47 \\ \hline
\end{tabular}
}}
\end{minipage}

\end{minipage}

\textbf{QLadder v.s. additional Visual Encoder.}
Recently, there has been works exploring the addition of extra visual encoders to achieve better visual representations, \textit{e.g.}, MOF~\cite{tong2024eyes}, which uses Dinov2~\cite{oquab2024dinov2} as a second visual encoder to enhance the grounding ability of MLLMs.
To explore the impact of adding QLadder and adding extra visual encoder, we conducted detailed experiments to directly compare Q-Ladder with the MoF method, which integrates DINOv2, under the LLaVA-v1.5 setting. The results are shown in Table~\ref{tab:additional_visual_encoder}.
Our experimental results show that both Q-Ladder and MoF performed well in visual grounding, achieving significant improvements on the MMVP and POPE benchmarks. However, MoF's performance declined on more comprehensive benchmarks like MMbench and OCR benchmarks like TextVQA. This decline is primarily due to MoF's reliance on DINOv2 for visual grounding, which, while enhancing grounding capabilities, weakened visual understanding, leading to poorer results on MMbench and TextVQA. Additionally, the integration of DINOv2 significantly increased the model's training time.
In contrast, Q-Ladder enhances both visual grounding and visual understanding through adaptive learning of distinguishing features. This dual improvement allows Q-Ladder to maintain or boost performance across a wide range of benchmarks, even when using a smaller dataset (over 2 million samples from Arcana). This is why Q-Ladder continues to achieve performance gains across various benchmarks, including comprehensive and OCR benchmarks.

\textbf{Impact of MM-LoRA and QLadder in MLLMs.}
To investigate the impact of MM-LoRA and QLadder in multimodal scenarios, we visualized the attention maps of Arcana with and without these modules in MM-Vet benchmark~\cite{yu2023mmvet}. The visualization results, shown in Fig.~\ref{fig:attention}, display the attention scores of generated tokens over the input sequence during the generation process. It can be seen that MLLM decoder initially focuses more on text tokens and gradually increases attention to visual tokens in the middle and subsequent layers. This indicates that visual and language information play different roles in MLLMs. The discussion about shallow-level attention maps, which also reflects this point, is provided in the Appendix.
Additionally, with MM-LoRA, we observe a significant increase in attention to visual tokens in the middle and subsequent layers, indicating that MM-LoRA helps prevent information confusion and promotes cooperation between different modalities. With the introduction of QLadder, the MLLM decoder shows increased attention to visual tokens across all layers. The highlighted regions of visual tokens further indicate that QLadder not only enhances the model's focus on visual tokens but also enriches the visual information, achieving optimal performance in multimodal tasks.

\begin{figure*}
	\centering
	\includegraphics[width=\linewidth]{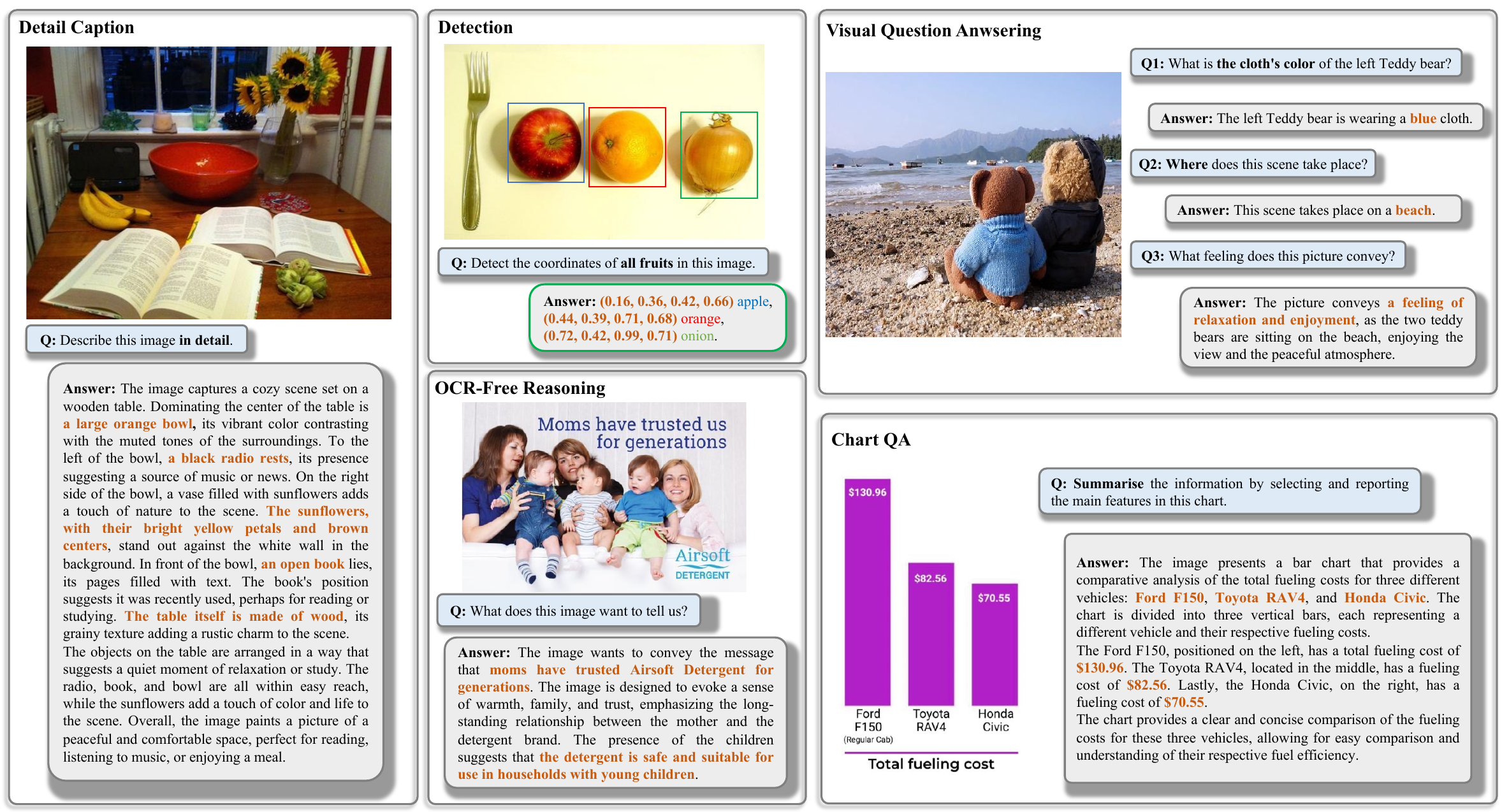}
        \vspace{-1.em}
	\caption{Examples of results generated by Arcana were sampled, focusing on tasks that test visual perception capabilities, such as detailed captions, detection, and OCR-reasoning. In the answers, all visual recognition-related responses are highlighted in \textcolor{orange}{orange}.}
	\label{fig:qualitative_results}
\vspace{-1.5em}
\end{figure*}

\textbf{Visualization results.}
To showcase Arcana's outstanding performance in visual perception, we visualized its performance across various types of multimodal tasks. As illustrated in Fig.~\ref{fig:qualitative_results}, visual perception information is highlighted in orange. In detailed description tasks, our model not only accurately identifies and describes low-level visual information such as colors and textures for each object in the image but also precisely recognizes and describes high-level visual information such as positions and relationships of each object. Moreover, detection tasks further demonstrate our model's effectiveness in visual recognition and localization. OCR-Free inference and chart-based question answering tasks not only exhibit our model's OCR recognition capabilities but also demonstrate its reasoning prowess. Visual question answering tasks showcase our model's excellent multi-turn dialogue capabilities on the foundation of precise identification.

In summary, Arcana utilizes a multimodal decoder to avoid information interference between different modalities. QLadder offers an innovative strategy for enhancing visual representations with limited data. By adding a small number of visual tokens, it significantly improves the performance of large multimodal language models. This finding is significant for the future of multimodal model, as it presents an effective approach to achieving notable performance improvements even with limited data resources. By combining these techniques, future multimodal models will handle complex tasks with greater flexibility and efficiency.

~\vspace{-1.5em}
\section{Conclusion}
\vspace{-1.1em}
In this paper, we introduce a new multimodal large language model, \textbf{Arcana}, which incorporates two novel techniques. Unlike current mainstream methods, Arcana employs MM-LoRA for a multimodal decoder, enabling more efficient information processing and integration across different modalities. MM-LoRA effectively combines data from various modalities without significantly increasing computational complexity, reducing information interference between modalities. Secondly, we present the QLadder structure, which demonstrates for the first time that with limited multimodal training data, retaining the capabilities of a pre-trained model and adding a small number of visual encoders can still enhance the performance of multimodal language models. This hierarchical structure progressively refines and enhances the expression of visual information, resulting in improved adaptability and generalization in multimodal tasks. With these two key techniques, Arcana not only excels in multimodal tasks but also shows potential for performance improvement even in data-constrained environments. Additionally, the severe lack of visual information in the image captions of open-source data limits the visual perception capabilities of multimodal large language models. To address this, we designed a data engine that uses diverse visual annotation models and large language models to generate captions rich in visual information.

{\small
\bibliographystyle{iclr2025_conference}
\bibliography{egbib.bib}

\begin{thebibliography}{58}
\providecommand{\natexlab}[1]{#1}
\providecommand{\url}[1]{\texttt{#1}}
\expandafter\ifx\csname urlstyle\endcsname\relax
  \providecommand{\doi}[1]{doi: #1}\else
  \providecommand{\doi}{doi: \begingroup \urlstyle{rm}\Url}\fi

\bibitem[Achiam et~al.(2023)Achiam, Adler, Agarwal, Ahmad, Akkaya, Aleman, Almeida, Altenschmidt, Altman, Anadkat, et~al.]{achiam2023gpt}
Josh Achiam, Steven Adler, Sandhini Agarwal, Lama Ahmad, Ilge Akkaya, Florencia~Leoni Aleman, Diogo Almeida, Janko Altenschmidt, Sam Altman, Shyamal Anadkat, et~al.
\newblock Gpt-4 technical report.
\newblock \emph{arXiv preprint arXiv:2303.08774}, 2023.

\bibitem[Alayrac et~al.(2022)Alayrac, Donahue, Luc, Miech, Barr, Hasson, Lenc, Mensch, Millican, Reynolds, et~al.]{alayrac2022flamingo}
Jean-Baptiste Alayrac, Jeff Donahue, Pauline Luc, Antoine Miech, Iain Barr, Yana Hasson, Karel Lenc, Arthur Mensch, Katherine Millican, Malcolm Reynolds, et~al.
\newblock Flamingo: a visual language model for few-shot learning.
\newblock In \emph{Advances in neural information processing systems}, pp.\  23716--23736, 2022.

\bibitem[Anil et~al.(2023)Anil, Dai, Firat, Johnson, Lepikhin, Passos, Shakeri, Taropa, Bailey, Chen, et~al.]{anil2023palm}
Rohan Anil, Andrew~M Dai, Orhan Firat, Melvin Johnson, Dmitry Lepikhin, Alexandre Passos, Siamak Shakeri, Emanuel Taropa, Paige Bailey, Zhifeng Chen, et~al.
\newblock Palm 2 technical report.
\newblock \emph{arXiv preprint arXiv:2305.10403}, 2023.

\bibitem[Bai et~al.(2023)Bai, Bai, Yang, Wang, Tan, Wang, Lin, Zhou, and Zhou]{bai2023qwen}
Jinze Bai, Shuai Bai, Shusheng Yang, Shijie Wang, Sinan Tan, Peng Wang, Junyang Lin, Chang Zhou, and Jingren Zhou.
\newblock Qwen-vl: A versatile vision-language model for understanding, localization, text reading, and beyond.
\newblock 2023.

\bibitem[Chen et~al.(2023{\natexlab{a}})Chen, Zhang, Zeng, Zhang, Zhu, and Zhao]{chen2023shikra}
Keqin Chen, Zhao Zhang, Weili Zeng, Richong Zhang, Feng Zhu, and Rui Zhao.
\newblock Shikra: Unleashing multimodal llm's referential dialogue magic.
\newblock \emph{arXiv preprint arXiv:2306.15195}, 2023{\natexlab{a}}.

\bibitem[Chen et~al.(2023{\natexlab{b}})Chen, Li, Dong, Zhang, He, Wang, Zhao, and Lin]{chen2023sharegpt4v}
Lin Chen, Jisong Li, Xiaoyi Dong, Pan Zhang, Conghui He, Jiaqi Wang, Feng Zhao, and Dahua Lin.
\newblock Sharegpt4v: Improving large multi-modal models with better captions.
\newblock \emph{arXiv preprint arXiv:2311.12793}, 2023{\natexlab{b}}.

\bibitem[Chen et~al.(2022)Chen, Wang, Changpinyo, Piergiovanni, Padlewski, Salz, Goodman, Grycner, Mustafa, Beyer, et~al.]{chen2022pali}
Xi~Chen, Xiao Wang, Soravit Changpinyo, AJ~Piergiovanni, Piotr Padlewski, Daniel Salz, Sebastian Goodman, Adam Grycner, Basil Mustafa, Lucas Beyer, et~al.
\newblock Pali: A jointly-scaled multilingual language-image model.
\newblock In \emph{The Eleventh International Conference on Learning Representations}, 2022.

\bibitem[Chiang et~al.(2023)Chiang, Li, Lin, Sheng, Wu, Zhang, Zheng, Zhuang, Zhuang, Gonzalez, et~al.]{chiang2023vicuna}
Wei-Lin Chiang, Zhuohan Li, Zi~Lin, Ying Sheng, Zhanghao Wu, Hao Zhang, Lianmin Zheng, Siyuan Zhuang, Yonghao Zhuang, Joseph~E Gonzalez, et~al.
\newblock Vicuna: An open-source chatbot impressing gpt-4 with 90\%* chatgpt quality.
\newblock \emph{See https://vicuna. lmsys. org (accessed 14 April 2023)}, 2\penalty0 (3):\penalty0 6, 2023.

\bibitem[Clark et~al.(2018)Clark, Cowhey, Etzioni, Khot, Sabharwal, Schoenick, and Tafjord]{clark2018ARC}
Peter Clark, Isaac Cowhey, Oren Etzioni, Tushar Khot, Ashish Sabharwal, Carissa Schoenick, and Oyvind Tafjord.
\newblock Think you have solved question answering? try arc, the ai2 reasoning challenge.
\newblock \emph{arXiv preprint arXiv:1803.05457}, 2018.

\bibitem[Dai et~al.(2024)Dai, Li, Li, Tiong, Zhao, Wang, Li, Fung, and Hoi]{dai2024instructblip}
Wenliang Dai, Junnan Li, Dongxu Li, Anthony Meng~Huat Tiong, Junqi Zhao, Weisheng Wang, Boyang Li, Pascale~N Fung, and Steven Hoi.
\newblock Instructblip: Towards general-purpose vision-language models with instruction tuning.
\newblock \emph{Advances in Neural Information Processing Systems}, 36, 2024.

\bibitem[Dong et~al.(2024)Dong, Zhang, Zang, Cao, Wang, Ouyang, Wei, Zhang, Duan, Cao, et~al.]{dong2024internlm}
Xiaoyi Dong, Pan Zhang, Yuhang Zang, Yuhang Cao, Bin Wang, Linke Ouyang, Xilin Wei, Songyang Zhang, Haodong Duan, Maosong Cao, et~al.
\newblock Internlm-xcomposer2: Mastering free-form text-image composition and comprehension in vision-language large model.
\newblock \emph{arXiv preprint arXiv:2401.16420}, 2024.

\bibitem[Driess et~al.(2023)Driess, Xia, Sajjadi, Lynch, Chowdhery, Ichter, Wahid, Tompson, Vuong, Yu, et~al.]{driess2023palm}
Danny Driess, Fei Xia, Mehdi~SM Sajjadi, Corey Lynch, Aakanksha Chowdhery, Brian Ichter, Ayzaan Wahid, Jonathan Tompson, Quan Vuong, Tianhe Yu, et~al.
\newblock Palm-e: An embodied multimodal language model.
\newblock In \emph{International Conference on Machine Learning}, pp.\  8469--8488. PMLR, 2023.

\bibitem[Gao et~al.(2023)Gao, Han, Zhang, Lin, Geng, Zhou, Zhang, Lu, He, Yue, et~al.]{gao2023llamaadapter2}
Peng Gao, Jiaming Han, Renrui Zhang, Ziyi Lin, Shijie Geng, Aojun Zhou, Wei Zhang, Pan Lu, Conghui He, Xiangyu Yue, et~al.
\newblock Llama-adapter v2: Parameter-efficient visual instruction model.
\newblock \emph{arXiv preprint arXiv:2304.15010}, 2023.

\bibitem[Goyal et~al.(2017)Goyal, Khot, Summers-Stay, Batra, and Parikh]{goyal2017vqav2}
Yash Goyal, Tejas Khot, Douglas Summers-Stay, Dhruv Batra, and Devi Parikh.
\newblock Making the v in vqa matter: Elevating the role of image understanding in visual question answering.
\newblock In \emph{Proceedings of the IEEE conference on computer vision and pattern recognition}, pp.\  6904--6913, 2017.

\bibitem[Hao et~al.(2024)Hao, Zhao, Chen, Sun, Chen, Zhang, Yao, Ding, and Wang]{hao2024fullanno}
Jing Hao, Yuxiang Zhao, Song Chen, Yanpeng Sun, Qiang Chen, Gang Zhang, Kun Yao, Errui Ding, and Jingdong Wang.
\newblock Fullanno: A data engine for enhancing image comprehension of mllms.
\newblock \emph{arXiv preprint arXiv:2409.13540}, 2024.

\bibitem[Hu et~al.(2021)Hu, Wallis, Allen-Zhu, Li, Wang, Wang, Chen, et~al.]{hu2021lora}
Edward~J Hu, Phillip Wallis, Zeyuan Allen-Zhu, Yuanzhi Li, Shean Wang, Lu~Wang, Weizhu Chen, et~al.
\newblock Lora: Low-rank adaptation of large language models.
\newblock In \emph{International Conference on Learning Representations}, 2021.

\bibitem[Hudson \& Manning(2019)Hudson and Manning]{hudson2019gqa}
Drew~A Hudson and Christopher~D Manning.
\newblock Gqa: A new dataset for real-world visual reasoning and compositional question answering.
\newblock In \emph{Proceedings of the IEEE conference on computer vision and pattern recognition}, pp.\  6700--6709, 2019.

\bibitem[Jiang et~al.(2023)Jiang, Sablayrolles, Mensch, Bamford, Chaplot, Casas, Bressand, Lengyel, Lample, Saulnier, et~al.]{jiang2023mistral}
Albert~Q Jiang, Alexandre Sablayrolles, Arthur Mensch, Chris Bamford, Devendra~Singh Chaplot, Diego de~las Casas, Florian Bressand, Gianna Lengyel, Guillaume Lample, Lucile Saulnier, et~al.
\newblock Mistral 7b.
\newblock \emph{arXiv preprint arXiv:2310.06825}, 2023.

\bibitem[Kazemzadeh et~al.(2014)Kazemzadeh, Ordonez, Matten, and Berg]{kazemzadeh2014referitgame}
Sahar Kazemzadeh, Vicente Ordonez, Mark Matten, and Tamara Berg.
\newblock Referitgame: Referring to objects in photographs of natural scenes.
\newblock In \emph{Proceedings of the 2014 conference on empirical methods in natural language processing (EMNLP)}, pp.\  787--798, 2014.

\bibitem[Kembhavi et~al.(2016)Kembhavi, Salvato, Kolve, Seo, Hajishirzi, and Farhadi]{ai2d}
Aniruddha Kembhavi, Mike Salvato, Eric Kolve, Minjoon Seo, Hannaneh Hajishirzi, and Ali Farhadi.
\newblock A diagram is worth a dozen images.
\newblock In \emph{European Conference on Computer Vision}, pp.\  235--251, 2016.

\bibitem[Krishna et~al.(2017)Krishna, Zhu, Groth, Johnson, Hata, Kravitz, Chen, Kalantidis, Li, Shamma, et~al.]{krishna2017visual}
Ranjay Krishna, Yuke Zhu, Oliver Groth, Justin Johnson, Kenji Hata, Joshua Kravitz, Stephanie Chen, Yannis Kalantidis, Li-Jia Li, David~A Shamma, et~al.
\newblock Visual genome: Connecting language and vision using crowdsourced dense image annotations.
\newblock \emph{International journal of computer vision}, 123:\penalty0 32--73, 2017.

\bibitem[Lauren{\c{c}}on et~al.(2024)Lauren{\c{c}}on, Saulnier, Tronchon, Bekman, Singh, Lozhkov, Wang, Karamcheti, Rush, Kiela, et~al.]{laurenccon2024obelics}
Hugo Lauren{\c{c}}on, Lucile Saulnier, L{\'e}o Tronchon, Stas Bekman, Amanpreet Singh, Anton Lozhkov, Thomas Wang, Siddharth Karamcheti, Alexander Rush, Douwe Kiela, et~al.
\newblock Obelics: An open web-scale filtered dataset of interleaved image-text documents.
\newblock In \emph{Advances in Neural Information Processing Systems}, volume~36, 2024.

\bibitem[Li et~al.(2023{\natexlab{a}})Li, Zhang, Chen, Wang, Yang, and Liu]{li2023otter}
Bo~Li, Yuanhan Zhang, Liangyu Chen, Jinghao Wang, Jingkang Yang, and Ziwei Liu.
\newblock Otter: A multi-modal model with in-context instruction tuning.
\newblock \emph{arXiv preprint arXiv:2305.03726}, 2023{\natexlab{a}}.

\bibitem[Li et~al.(2023{\natexlab{b}})Li, Wang, Wang, Ge, Ge, and Shan]{li2023seed-bench}
Bohao Li, Rui Wang, Guangzhi Wang, Yuying Ge, Yixiao Ge, and Ying Shan.
\newblock Seed-bench: Benchmarking multimodal llms with generative comprehension.
\newblock \emph{arXiv preprint arXiv:2307.16125}, 2023{\natexlab{b}}.

\bibitem[Li et~al.(2022)Li, Li, Xiong, and Hoi]{li2022blip}
Junnan Li, Dongxu Li, Caiming Xiong, and Steven Hoi.
\newblock Blip: Bootstrapping language-image pre-training for unified vision-language understanding and generation.
\newblock In \emph{International conference on machine learning}, pp.\  12888--12900. PMLR, 2022.

\bibitem[Li et~al.(2023{\natexlab{c}})Li, Du, Zhou, Wang, Zhao, and Wen]{li2023pope}
Yifan Li, Yifan Du, Kun Zhou, Jinpeng Wang, Xin Zhao, and Ji-Rong Wen.
\newblock Evaluating object hallucination in large vision-language models.
\newblock In \emph{The 2023 Conference on Empirical Methods in Natural Language Processing}, 2023{\natexlab{c}}.

\bibitem[Liu et~al.(2023{\natexlab{a}})Liu, Li, Li, and Lee]{liu2023llava1.5}
Haotian Liu, Chunyuan Li, Yuheng Li, and Yong~Jae Lee.
\newblock Improved baselines with visual instruction tuning.
\newblock \emph{arXiv preprint arXiv:2310.03744}, 2023{\natexlab{a}}.

\bibitem[Liu et~al.(2024)Liu, Li, Wu, and Lee]{llava}
Haotian Liu, Chunyuan Li, Qingyang Wu, and Yong~Jae Lee.
\newblock Visual instruction tuning.
\newblock \emph{Advances in neural information processing systems}, 36, 2024.

\bibitem[Liu et~al.(2023{\natexlab{b}})Liu, Duan, Zhang, Li, Zhang, Zhao, Yuan, Wang, He, Liu, et~al.]{liu2023mmbench}
Yuan Liu, Haodong Duan, Yuanhan Zhang, Bo~Li, Songyang Zhang, Wangbo Zhao, Yike Yuan, Jiaqi Wang, Conghui He, Ziwei Liu, et~al.
\newblock Mmbench: Is your multi-modal model an all-around player?
\newblock \emph{arXiv preprint arXiv:2307.06281}, 2023{\natexlab{b}}.

\bibitem[Liu et~al.(2022)Liu, Mao, Wu, Feichtenhofer, Darrell, and Xie]{liu2022convnet}
Zhuang Liu, Hanzi Mao, Chao-Yuan Wu, Christoph Feichtenhofer, Trevor Darrell, and Saining Xie.
\newblock A convnet for the 2020s.
\newblock In \emph{Proceedings of the IEEE conference on computer vision and pattern recognition}, pp.\  11976--11986, 2022.

\bibitem[Lu et~al.(2022)Lu, Mishra, Xia, Qiu, Chang, Zhu, Tafjord, Clark, and Kalyan]{scienceQA}
Pan Lu, Swaroop Mishra, Tanglin Xia, Liang Qiu, Kai-Wei Chang, Song-Chun Zhu, Oyvind Tafjord, Peter Clark, and Ashwin Kalyan.
\newblock Learn to explain: Multimodal reasoning via thought chains for science question answering.
\newblock In \emph{Advances in Neural Information Processing Systems}, pp.\  2507--2521, 2022.

\bibitem[Luo et~al.(2024)Luo, Zhou, Zhang, Zheng, Sun, and Ji]{luo2024feast}
Gen Luo, Yiyi Zhou, Yuxin Zhang, Xiawu Zheng, Xiaoshuai Sun, and Rongrong Ji.
\newblock Feast your eyes: Mixture-of-resolution adaptation for multimodal large language models.
\newblock \emph{arXiv preprint arXiv:2403.03003}, 2024.

\bibitem[Mao et~al.(2016)Mao, Huang, Toshev, Camburu, Yuille, and Murphy]{mao2016generation}
Junhua Mao, Jonathan Huang, Alexander Toshev, Oana Camburu, Alan~L Yuille, and Kevin Murphy.
\newblock Generation and comprehension of unambiguous object descriptions.
\newblock In \emph{Proceedings of the IEEE conference on computer vision and pattern recognition}, pp.\  11--20, 2016.

\bibitem[Marino et~al.(2019)Marino, Rastegari, Farhadi, and Mottaghi]{marino2019ok}
Kenneth Marino, Mohammad Rastegari, Ali Farhadi, and Roozbeh Mottaghi.
\newblock Ok-vqa: A visual question answering benchmark requiring external knowledge.
\newblock In \emph{Proceedings of the IEEE conference on computer vision and pattern recognition}, pp.\  3195--3204, 2019.

\bibitem[Meng et~al.(2021)Meng, Chen, Fan, Zeng, Li, Yuan, Sun, and Wang]{meng2021conditional}
Depu Meng, Xiaokang Chen, Zejia Fan, Gang Zeng, Houqiang Li, Yuhui Yuan, Lei Sun, and Jingdong Wang.
\newblock Conditional detr for fast training convergence.
\newblock In \emph{Proceedings of the IEEE/CVF international conference on computer vision}, pp.\  3651--3660, 2021.

\bibitem[Mishra et~al.(2019)Mishra, Shekhar, Singh, and Chakraborty]{mishra2019ocr}
Anand Mishra, Shashank Shekhar, Ajeet~Kumar Singh, and Anirban Chakraborty.
\newblock Ocr-vqa: Visual question answering by reading text in images.
\newblock In \emph{2019 international conference on document analysis and recognition (ICDAR)}, pp.\  947--952. IEEE, 2019.

\bibitem[Oquab et~al.(2024)Oquab, Darcet, Moutakanni, Vo, Szafraniec, Khalidov, Fernandez, Haziza, Massa, El-Nouby, et~al.]{oquab2024dinov2}
Maxime Oquab, Timoth{\'e}e Darcet, Th{\'e}o Moutakanni, Huy Vo, Marc Szafraniec, Vasil Khalidov, Pierre Fernandez, Daniel Haziza, Francisco Massa, Alaaeldin El-Nouby, et~al.
\newblock Dinov2: Learning robust visual features without supervision.
\newblock \emph{Transactions on Machine Learning Research Journal}, pp.\  1--31, 2024.

\bibitem[Radford et~al.(2021)Radford, Kim, Hallacy, Ramesh, Goh, Agarwal, Sastry, Askell, Mishkin, Clark, et~al.]{radford2021learning}
Alec Radford, Jong~Wook Kim, Chris Hallacy, Aditya Ramesh, Gabriel Goh, Sandhini Agarwal, Girish Sastry, Amanda Askell, Pamela Mishkin, Jack Clark, et~al.
\newblock Learning transferable visual models from natural language supervision.
\newblock In \emph{International conference on machine learning}, pp.\  8748--8763. PMLR, 2021.

\bibitem[Schwenk et~al.(2022)Schwenk, Khandelwal, Clark, Marino, and Mottaghi]{schwenk2022okvqa}
Dustin Schwenk, Apoorv Khandelwal, Christopher Clark, Kenneth Marino, and Roozbeh Mottaghi.
\newblock A-okvqa: A benchmark for visual question answering using world knowledge.
\newblock In \emph{European Conference on Computer Vision}, pp.\  146--162. Springer, 2022.

\bibitem[Share{GPT}(2023)]{sharegpt}
Share{GPT}.
\newblock \url{https://sharegpt.com/}, 2023.

\bibitem[Sidorov et~al.(2020)Sidorov, Hu, Rohrbach, and Singh]{sidorov2020textcaps}
Oleksii Sidorov, Ronghang Hu, Marcus Rohrbach, and Amanpreet Singh.
\newblock Textcaps: a dataset for image captioning with reading comprehension.
\newblock In \emph{European Conference on Computer Vision}. Springer, 2020.

\bibitem[Singh et~al.(2019)Singh, Natarajan, Shah, Jiang, Chen, Batra, Parikh, and Rohrbach]{textvqa}
Amanpreet Singh, Vivek Natarajan, Meet Shah, Yu~Jiang, Xinlei Chen, Dhruv Batra, Devi Parikh, and Marcus Rohrbach.
\newblock Towards vqa models that can read.
\newblock In \emph{Proceedings of the IEEE conference on computer vision and pattern recognition}, pp.\  8317--8326, 2019.

\bibitem[Suzgun et~al.(2023)Suzgun, Scales, Sch{\"a}rli, Gehrmann, Tay, Chung, Chowdhery, Le, Chi, Zhou, et~al.]{suzgun2023bbh}
Mirac Suzgun, Nathan Scales, Nathanael Sch{\"a}rli, Sebastian Gehrmann, Yi~Tay, Hyung~Won Chung, Aakanksha Chowdhery, Quoc Le, Ed~Chi, Denny Zhou, et~al.
\newblock Challenging big-bench tasks and whether chain-of-thought can solve them.
\newblock In \emph{Findings of the Association for Computational Linguistics: ACL 2023}, pp.\  13003--13051, 2023.

\bibitem[Tong et~al.(2024)Tong, Liu, Zhai, Ma, LeCun, and Xie]{tong2024eyes}
Shengbang Tong, Zhuang Liu, Yuexiang Zhai, Yi~Ma, Yann LeCun, and Saining Xie.
\newblock Eyes wide shut? exploring the visual shortcomings of multimodal llms.
\newblock \emph{arXiv preprint arXiv:2401.06209}, 2024.

\bibitem[Touvron et~al.(2023{\natexlab{a}})Touvron, Lavril, Izacard, Martinet, Lachaux, Lacroix, Rozi{\`e}re, Goyal, Hambro, Azhar, et~al.]{touvron2023llama}
Hugo Touvron, Thibaut Lavril, Gautier Izacard, Xavier Martinet, Marie-Anne Lachaux, Timoth{\'e}e Lacroix, Baptiste Rozi{\`e}re, Naman Goyal, Eric Hambro, Faisal Azhar, et~al.
\newblock Llama: Open and efficient foundation language models.
\newblock \emph{arXiv preprint arXiv:2302.13971}, 2023{\natexlab{a}}.

\bibitem[Touvron et~al.(2023{\natexlab{b}})Touvron, Martin, Stone, Albert, Almahairi, Babaei, Bashlykov, Batra, Bhargava, Bhosale, et~al.]{touvron2023llama2}
Hugo Touvron, Louis Martin, Kevin Stone, Peter Albert, Amjad Almahairi, Yasmine Babaei, Nikolay Bashlykov, Soumya Batra, Prajjwal Bhargava, Shruti Bhosale, et~al.
\newblock Llama 2: Open foundation and fine-tuned chat models.
\newblock \emph{arXiv preprint arXiv:2307.09288}, 2023{\natexlab{b}}.

\bibitem[Wang et~al.(2022)Wang, Yang, Hu, Li, Lin, Gan, Liu, Liu, and Wang]{wang2022git}
Jianfeng Wang, Zhengyuan Yang, Xiaowei Hu, Linjie Li, Kevin Lin, Zhe Gan, Zicheng Liu, Ce~Liu, and Lijuan Wang.
\newblock Git: A generative image-to-text transformer for vision and language.
\newblock \emph{Transactions on Machine Learning Research}, 2022.

\bibitem[Wang et~al.(2023)Wang, Lv, Yu, Hong, Qi, Wang, Ji, Yang, Zhao, Song, et~al.]{wang2023cogvlm}
Weihan Wang, Qingsong Lv, Wenmeng Yu, Wenyi Hong, Ji~Qi, Yan Wang, Junhui Ji, Zhuoyi Yang, Lei Zhao, Xixuan Song, et~al.
\newblock Cogvlm: Visual expert for pretrained language models.
\newblock \emph{arXiv preprint arXiv:2311.03079}, 2023.

\bibitem[Xu et~al.(2023)Xu, Sun, Zheng, Geng, Zhao, Feng, Tao, and Jiang]{xu2023wizardlm}
Can Xu, Qingfeng Sun, Kai Zheng, Xiubo Geng, Pu~Zhao, Jiazhan Feng, Chongyang Tao, and Daxin Jiang.
\newblock Wizardlm: Empowering large language models to follow complex instructions.
\newblock \emph{arXiv preprint arXiv:2304.12244}, 2023.

\bibitem[Xu et~al.(2024)Xu, Yao, Guo, Cui, Ni, Ge, Chua, Liu, Sun, and Huang]{xu2024llava-uhd}
Ruyi Xu, Yuan Yao, Zonghao Guo, Junbo Cui, Zanlin Ni, Chunjiang Ge, Tat-Seng Chua, Zhiyuan Liu, Maosong Sun, and Gao Huang.
\newblock Llava-uhd: an lmm perceiving any aspect ratio and high-resolution images.
\newblock \emph{arXiv preprint arXiv:2403.11703}, 2024.

\bibitem[Ye et~al.(2023{\natexlab{a}})Ye, Xu, Xu, Ye, Yan, Zhou, Wang, Hu, Shi, Shi, et~al.]{ye2023mplug}
Qinghao Ye, Haiyang Xu, Guohai Xu, Jiabo Ye, Ming Yan, Yiyang Zhou, Junyang Wang, Anwen Hu, Pengcheng Shi, Yaya Shi, et~al.
\newblock mplug-owl: Modularization empowers large language models with multimodality.
\newblock \emph{arXiv preprint arXiv:2304.14178}, 2023{\natexlab{a}}.

\bibitem[Ye et~al.(2023{\natexlab{b}})Ye, Xu, Ye, Yan, Liu, Qian, Zhang, Huang, and Zhou]{ye2023mplug2}
Qinghao Ye, Haiyang Xu, Jiabo Ye, Ming Yan, Haowei Liu, Qi~Qian, Ji~Zhang, Fei Huang, and Jingren Zhou.
\newblock mplug-owl2: Revolutionizing multi-modal large language model with modality collaboration.
\newblock \emph{arXiv preprint arXiv:2311.04257}, 2023{\natexlab{b}}.

\bibitem[Yu et~al.(2023)Yu, Yang, Li, Wang, Lin, Liu, Wang, and Wang]{yu2023mmvet}
Weihao Yu, Zhengyuan Yang, Linjie Li, Jianfeng Wang, Kevin Lin, Zicheng Liu, Xinchao Wang, and Lijuan Wang.
\newblock Mm-vet: Evaluating large multimodal models for integrated capabilities.
\newblock \emph{arXiv preprint arXiv:2308.02490}, 2023.

\bibitem[Zhang et~al.(2023)Zhang, Wang, Cao, Xu, Ouyang, Zhao, Ding, Zhang, Duan, Yan, et~al.]{zhang2023xcomposer}
Pan Zhang, Xiaoyi Dong~Bin Wang, Yuhang Cao, Chao Xu, Linke Ouyang, Zhiyuan Zhao, Shuangrui Ding, Songyang Zhang, Haodong Duan, Hang Yan, et~al.
\newblock Internlm-xcomposer: A vision-language large model for advanced text-image comprehension and composition.
\newblock \emph{arXiv preprint arXiv:2309.15112}, 2023.

\bibitem[Zheng et~al.(2024)Zheng, Chiang, Sheng, Zhuang, Wu, Zhuang, Lin, Li, Li, Xing, et~al.]{zheng2024judging}
Lianmin Zheng, Wei-Lin Chiang, Ying Sheng, Siyuan Zhuang, Zhanghao Wu, Yonghao Zhuang, Zi~Lin, Zhuohan Li, Dacheng Li, Eric Xing, et~al.
\newblock Judging llm-as-a-judge with mt-bench and chatbot arena.
\newblock In \emph{Advances in Neural Information Processing Systems}, 2024.

\bibitem[Zhong et~al.(2023)Zhong, Cui, Guo, Liang, Lu, Wang, Saied, Chen, and Duan]{zhong2023agieval}
Wanjun Zhong, Ruixiang Cui, Yiduo Guo, Yaobo Liang, Shuai Lu, Yanlin Wang, Amin Saied, Weizhu Chen, and Nan Duan.
\newblock Agieval: A human-centric benchmark for evaluating foundation models.
\newblock \emph{arXiv preprint arXiv:2304.06364}, 2023.

\bibitem[Zhu et~al.(2023)Zhu, Chen, Shen, Li, and Elhoseiny]{zhu2023minigpt}
Deyao Zhu, Jun Chen, Xiaoqian Shen, Xiang Li, and Mohamed Elhoseiny.
\newblock Minigpt-4: Enhancing vision-language understanding with advanced large language models.
\newblock In \emph{The Twelfth International Conference on Learning Representations}, 2023.

\bibitem[Zhu et~al.(2020)Zhu, Su, Lu, Li, Wang, and Dai]{zhu2020deformable}
Xizhou Zhu, Weijie Su, Lewei Lu, Bin Li, Xiaogang Wang, and Jifeng Dai.
\newblock Deformable detr: Deformable transformers for end-to-end object detection.
\newblock In \emph{International Conference on Learning Representations}, 2020.

\end{thebibliography}
}

\end{document}


\maketitle

~\appendix
\vspace{-5.em}
\section{Appendix}
\subsection{Detailed Evaluation Results.}
\textbf{POPE.}
We conduct the hallucination evaluation using POPE~\cite{li2023pope}, the results are shown in Table. From the results in the Table~\ref{tab:pope}, we can find Arcana achieves higher F1 scores on the popular and adversarial split, showing the robustness of our model in terms of object hallucination compared to other MLLMs.

\textbf{MMBench.}
MMBench~\cite{liu2023mmbench} is used to evaluate the model’s ability of Perception and Reasoning. The detail results for various MLLMs are presented in Table~\ref{tab:mmbench}.

\subsection{More visualization results.}
\begin{figure*}
	\centering
	\includegraphics[width=\linewidth]{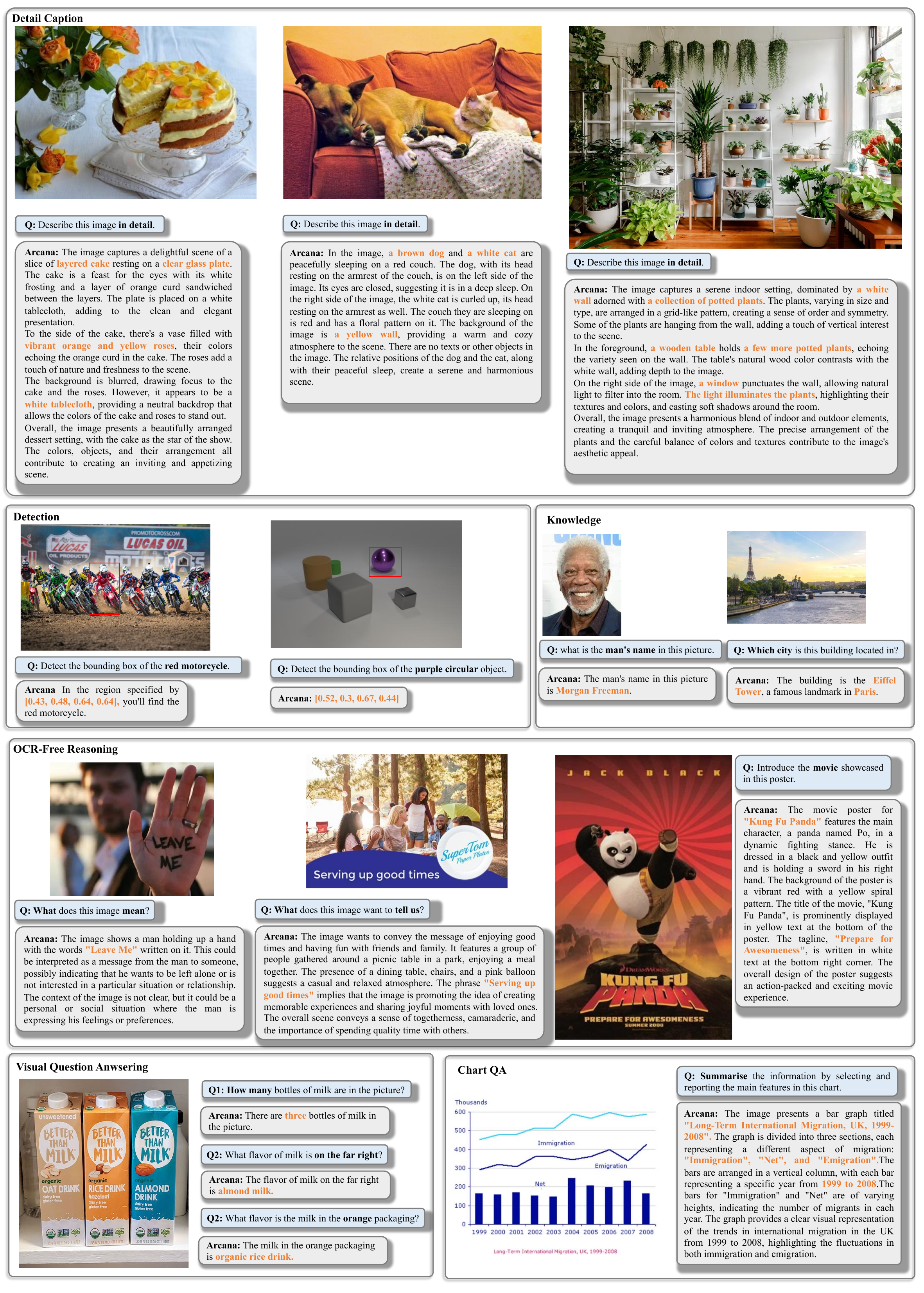}
	\caption{More qualitative results. Main feature in answer is highlight in \textbf{\textcolor{orange}{orange}}.}
	\label{fig:qualitative_results_supply}
\end{figure*}

\begin{figure*}
	\centering
	\includegraphics[width=\linewidth]{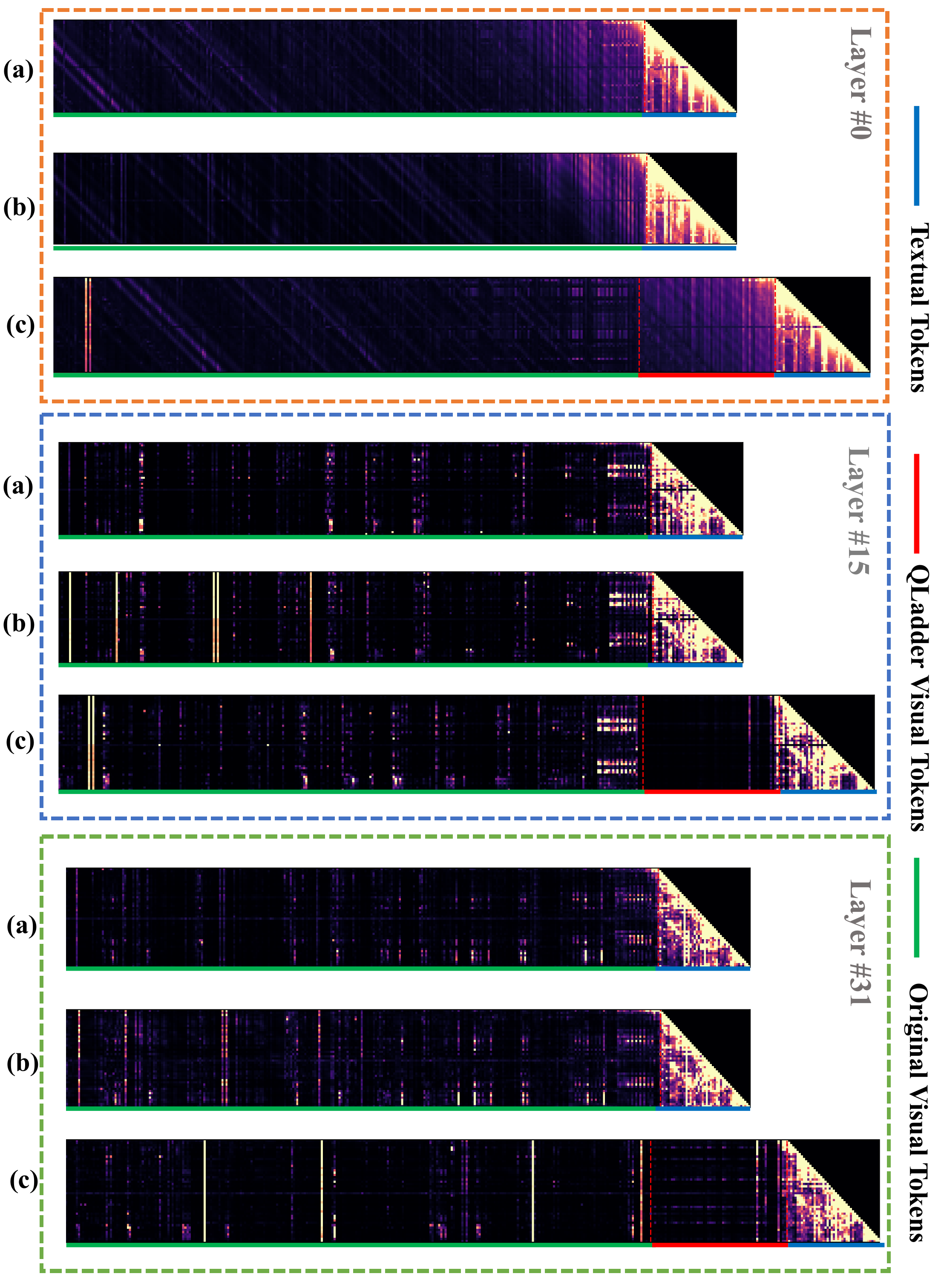}
	\caption{\textbf{Visualization of attention maps}. We compare the attention maps in different layer of LLM between different composition, include \textbf{(a)} Baseline, \textbf{(b)}Baseline+MM-LoRA, and \textbf{(c)} Baseline+MM LoRA+QLadder. Higher brightness indicates higher attention values, with the x-axis representing all tokens, and the y-axis containing only text tokens.}
	\label{fig:attention_comparision_supply}
\end{figure*}

To demonstrate the effectiveness and generalization ability of Arcana, we provide more qualitative results in Fig.~\ref{fig:qualitative_results_supply}. We visualize its performance across various types of multimodal tasks, including Detail Caption, Detection, Knowledge, OCR-Free Reasoning, Visual Question Answering and ChartQA. To investigate the impact of MM-LoRA and QLadder in multimodal scenarios, we visualized the attention maps of different layers with and without these modules in Fig.~\ref{fig:attention_comparision_supply}.

\subsection{Broader Impact}
\label{subsec:broader_impact}
This paper present Arcana, which target at improving the visual understanding capability for boosting the vision-language models.
To achieve this goal, Arcana conducts a series of explorations into visual learning within the model structure. On one hand, Arcana demonstrates that decoupling the learning of visual and language representation within the LLM is beneficial for avoiding information confusion while preserving the uniqueness of each modality, and based on this, proposes MM-LoRA.
On the other hand, Arcana asserts that under limited training data, it is important to retain the pre-trained image encoder's capabilities and introduces QLadder, which incorporates a small number of visual tokens to enhance the model's learning and representation abilities for visual information.
Extensive experiments demonstrate the effectiveness and generalization ability of Arcana.

The positive societal impacts of the work include:
\begin{itemize}
    \item \textbf{Improved Human-Machine Interaction}: Enhanced visual perception in multimodal models can lead to more intuitive and effective human-machine interactions. This could improve applications such as virtual assistants, customer service bots, and educational tools, making them more responsive and capable of understanding complex visual contexts.

    \item \textbf{Advancements in AI Research}: The Arcana model's innovative architecture and data handling approaches could stimulate further research in the AI community, leading to new breakthroughs and applications in various fields, from healthcare to autonomous vehicles, where precise visual perception is crucial.

    \item \textbf{Better Performance in Real-World Applications}: By addressing the deficiencies in low-level and high-level visual perception, Arcana can improve performance in practical applications like object detection in surveillance, quality control in manufacturing, and detailed image analysis in medical diagnostics.
\end{itemize}

The negative societal impacts may include:
\begin{itemize}
    \item \textbf{Privacy Concerns}: Enhanced visual perception capabilities may lead to more invasive surveillance technologies. The ability to detect and interpret small objects and detailed visual information could be misused to infringe on individuals' privacy, leading to unauthorized tracking and monitoring.

    \item \textbf{Security Risks}: Advanced visual perception models could be exploited for malicious purposes, such as by enhancing the capabilities of autonomous weapons or by improving the precision of surveillance systems used by authoritarian regimes to suppress dissent.

    \item \textbf{Dependence on Technology}: Increasing reliance on advanced AI for visual tasks may lead to a decrease in human skills and awareness in certain fields. Over-dependence on such technology without proper human oversight could have negative implications for critical decision-making processes.
\end{itemize}

\begin{table*}[t]
\centering
\setlength\tabcolsep{4pt}
\renewcommand\arraystretch{1.3}
\setlength{\tabcolsep}{2mm}{
\caption{Object hallucination benchmark using POPE evaluation pipeline. "Yes" signifies the likelihood of the model producing a positive response.}
\label{tab:pope}
\resizebox{\textwidth}{!}{%
\begin{tabular}{l|c|cccccc}
\toprule
Datasets & Metrics & \textbf{Arcana (Ours)} & mPLUG-Owl2~\cite{ye2023mplug2} & LLaVA-v1.5~\cite{liu2023llava1.5} & Shikra~\cite{chen2023shikra} & InstructBLIP~\cite{dai2024instructblip} &MiniGPT-4~\cite{zhu2023minigpt}  \\
\midrule
\multirow{5}{*}{Random} & Accuracy ($\uparrow$) & 88.87 & 88.28 & 88.38 & 86.90 & 88.57 & 79.67 \\
& Precision ($\uparrow$)  & 96.59 & 94.34 & 96.56 & 94.40 & 84.09 & 78.24 \\
& Recall ($\uparrow$)  & 81.27 & 82.20 & 80.33 & 79.27 & 95.13 & 82.20 \\
& F1-Score ($\uparrow$) & \underline{88.27} & 87.85 & 87.70 & 86.19 & \textbf{89.27} & 80.17 \\
& Yes ($\rightarrow$ 50\%) & 43.37 &44.91 & 42.89 & 43.26 & 56.57 & 52.53 \\
\midrule
\multirow{5}{*}{Popular} & Accuracy ($\uparrow$) & 88.07 & 86.20 & 87.67 & 83.97 & 82.77 & 69.73\\
& Precision ($\uparrow$)  & 94.06 & 89.46 & 94.14 & 87.55 & 76.27 & 65.86 \\
& Recall ($\uparrow$)  & 81.27 & 82.06 & 80.33 & 79.20 & 95.13 & 81.93 \\
& F1-Score ($\uparrow$) & \textbf{87.20} & 85.60 & \underline{86.69}  & 83.16 & 84.66 & 73.02 \\
& Yes ($\rightarrow$ 50\%) & 43.20 & 45.86 & 42.67  & 45.23 & 62.37 & 62.20 \\
\midrule
\multirow{5}{*}{Adversarial} & Accuracy ($\uparrow$) & 86.57 & 84.12 & 85.23 & 83.10 & 72.10 & 65.17\\
& Precision ($\uparrow$)  & 90.90 & 85.54 & 89.06 & 85.60 & 65.13 & 61.19 \\
& Recall ($\uparrow$)  & 81.27 & 82.13 & 80.33 & 79.60 & 95.13 & 82.93 \\
& F1-Score ($\uparrow$) & \textbf{85.81} & 83.80 & \underline{84.47} & 82.49 & 77.32 & 70.42 \\
& Yes ($\rightarrow$ 50\%) & 44.70 & 48.00 & 45.10  & 46.50 & 73.03 & 67.77 \\
\bottomrule
\end{tabular}
}}
\end{table*}

\begin{table*}[h]
\centering
\setlength\tabcolsep{4pt}
\renewcommand\arraystretch{1.3}
\setlength{\tabcolsep}{2mm}{
\caption{CircularEval multi-choice accuracy results on MMBench~\cite{liu2023mmbench} dev set.
We adopt the following abbreviations: LR for Logical
Reasoning; AR for Attribute Reasoning; RR for Relation Reasoning; FP-C for Fine-grained Perception (Cross Instance); FP-S for Finegrained
Perception (Single Instance); CP for Coarse Perception.}
\label{tab:mmbench}
\resizebox{\textwidth}{!}{%
\begin{tabular}{l|cc|ccccccc}
\toprule
Method & Language Model & Vision Model & Overall & LR & AR & RR & FP-S & FP-C & CP \\
\midrule
MiniGPT-4~\cite{zhu2023minigpt}& Vicuna-7B & EVA-G & 12.0 & 13.6 & 32.9 & 8.9 & 28.8 & 11.2 & 28.3 \\
InstructBLIP~\cite{dai2024instructblip} & Vicuna-7B & EVA-G & 33.9 & 21.6 & 47.4 & 22.5 & 33.0 & 24.4 & 41.1 \\
LLaMA-Adapter-v2~\cite{gao2023llamaadapter2} & LLaMa-7B & CLIP ViT-L/14 & 38.9 & 7.4 & 45.3 & 19.2 & 45.0 & 32.0 & 54.0 \\
LLaVA~\cite{llava}& Vicuna-7B & CLIP ViT L/14 & 36.2 & 15.9 & 53.6 & 28.6 & 41.8 & 20.0 & 40.4 \\
Shikra~\cite{chen2023shikra} & Vicuna-7B & CLIP ViT-L/14 & 60.2 & 33.5 & \underline{69.6} & 53.1 & 61.8 & 50.4 & 71.7 \\
LLaVA-v1.5~\cite{liu2023llava1.5} & Vicuna-7B & CLIP ViT-L/14 & 64.3 & 33.1 & 69.3 & 57.4 & 68.9 & 54.5 & 76.4 \\
mPLUG-Owl2~\cite{ye2023mplug2} & LLaMA2-7B & CLIP ViT-L/14 & 65.4 & 29.2 & \textbf{69.7} & 61.7 & 67.0 & \textbf{60.0} & 79.5 \\
\midrule
\textbf{Arcana (Ours)} & Vicuna-7B & CLIP ViT-L/14 & \textbf{67.4} & \textbf{34.7} & 69.3 & \textbf{62.6} & \textbf{69.6} & \underline{58.7} & \textbf{83.1} \\
\bottomrule
\end{tabular}
}}
\end{table*}

In summary, while the Arcana model holds promise for significant advancements and positive contributions to society, it is crucial to address the associated risks through responsible development, deployment, and regulation to mitigate potential negative impacts.

\subsection{Limitations and Future Work}
\label{subsec:limitation_futurework} 
The previous experiments have demonstrated the effectiveness of Arcana. Although the multimodal decoder has proven effective, giving each modality its own learning space significantly increases the number of parameters. While MM-LoRA only adds a small number of parameters to achieve a multimodal decoder, the independent LoRA parameters for different modalities cannot be merged into the LLMs' weights, thereby increasing inference costs. The introduction of QLadder enhances visual representation capabilities, but it comes at the cost of adding visual tokens, which also increases inference costs. Additionally, compared to existing state-of-the-art methods, we used only about 2M training data, limiting Arcana's performance. 

To further unlock Arcana's potential, we will design a more efficient multimodal decoder that improves performance while reducing inference costs. We will also focus on designing a more efficient visual encoder that uses fewer visual tokens to represent visual features, enhancing training efficiency and reducing inference costs. Finally, we plan to leverage our data engine to annotate more high-quality caption data to fully unleash Arcana's potential.

{\small
\bibliographystyle{iclr2025_conference}
\bibliography{egbib.bib}
}

